\documentclass{zlab}

\usepackage[T1]{fontenc}
\usepackage{tgpagella}
\usepackage{mathpazo}
\usepackage{inconsolata}
\usepackage{xspace}
\usepackage{graphicx,amsmath,amssymb,hyperref}
\usepackage[authoryear,round]{natbib}
\usepackage[table]{xcolor}
\usepackage{booktabs}
\usepackage{multirow}
\usepackage{array}
\usepackage{float}
\usepackage{tcolorbox}
\tcbuselibrary{skins,breakable}
\usepackage{capt-of}
\usepackage{fvextra}
\usepackage{tabularx}
\usepackage{wrapfig}
\usepackage{orcidlink}
\usepackage{etoc}

\graphicspath{{./}{../}{../figures/}{figures/}}

\newcommand{\dataengine}{\textbf{VisionFoundry}\xspace}
\newcommand{\datasetname}{\textbf{VisionFoundry-10K}\xspace}
\newcommand{\datasetnamenosym}{VisionFoundry-10K\xspace}
\newcommand{\synthcol}{gray!8}
\renewcommand{\paragraph}[1]{\vspace{1.25mm}\noindent\textbf{#1}}

\RecustomVerbatimEnvironment{verbatim}{Verbatim}{
  breaklines=true,
  breakanywhere=true,
  breaksymbolleft={},
  breaksymbolright={},
  fontsize=\small
}

\newcolumntype{Y}{>{\centering\arraybackslash}X}
\definecolor{prompttitle}{HTML}{404040}
\newtcolorbox{promptstylebox}[1]{
  breakable,
  enhanced,
  colback=gray!12,
  colframe=gray!55,
  colbacktitle=prompttitle,
  coltitle=white,
  fonttitle=\sffamily\fontseries{bx}\selectfont,
  fontupper=\small\sffamily,
  title={#1},
  boxrule=0.9pt,
  left=10pt, right=10pt, top=8pt, bottom=8pt,
  toptitle=5pt,
  bottomtitle=5pt,
  lefttitle=10pt,
  boxsep=0pt,
  sharp corners
}

\hypersetup{
  colorlinks=true,
  urlcolor=magenta,
  linkcolor=red,
  citecolor=link,
  filecolor=black,
  pdfborder={0 0 0}
}

\definecolor{link}{HTML}{0063BE}
\definecolor{appendixaccent}{HTML}{FF0000}
\providecommand{\equationname}{Equation}
\providecommand{\sectionname}{Section}

\newcommand{\figref}[1]{%
  \figurename~\hyperref[#1]{\textcolor{link}{\ref*{#1}}}%
}
\newcommand{\tabref}[1]{%
  \tablename~\hyperref[#1]{\textcolor{link}{\ref*{#1}}}%
}
\newcommand{\eqrefc}[1]{%
  \equationname~\hyperref[#1]{\textcolor{link}{\ref*{#1}}}%
}
\newcommand{\secrefc}[1]{%
  \sectionname~\hyperref[#1]{\textcolor{link}{\ref*{#1}}}%
}

\makeatletter
\newcommand{\appendixsectiontocline}[1]{%
  \begingroup\hypersetup{linkcolor=black}%
  \@dottedtocline{1}{0em}{2.2em}{\hyperref[#1]{\textbf{\ref*{#1}\hspace{0.55em}\nameref*{#1}}}}{\textbf{\hyperref[#1]{\pageref*{#1}}}}%
  \endgroup
}
\newcommand{\appendixsubsectiontocline}[1]{%
  \begingroup\hypersetup{linkcolor=black}%
  \@dottedtocline{2}{1.7em}{2.8em}{\hyperref[#1]{\ref*{#1}\hspace{0.55em}\nameref*{#1}}}{\hyperref[#1]{\pageref*{#1}}}%
  \endgroup
}
\makeatother

\newcommand{\github}{\raisebox{-1.3pt}{\includegraphics[height=1.05em]{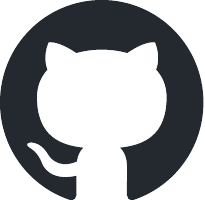}}\xspace}
\newcommand{\worldwideweb}{\raisebox{-1.3pt}{\includegraphics[height=1.05em]{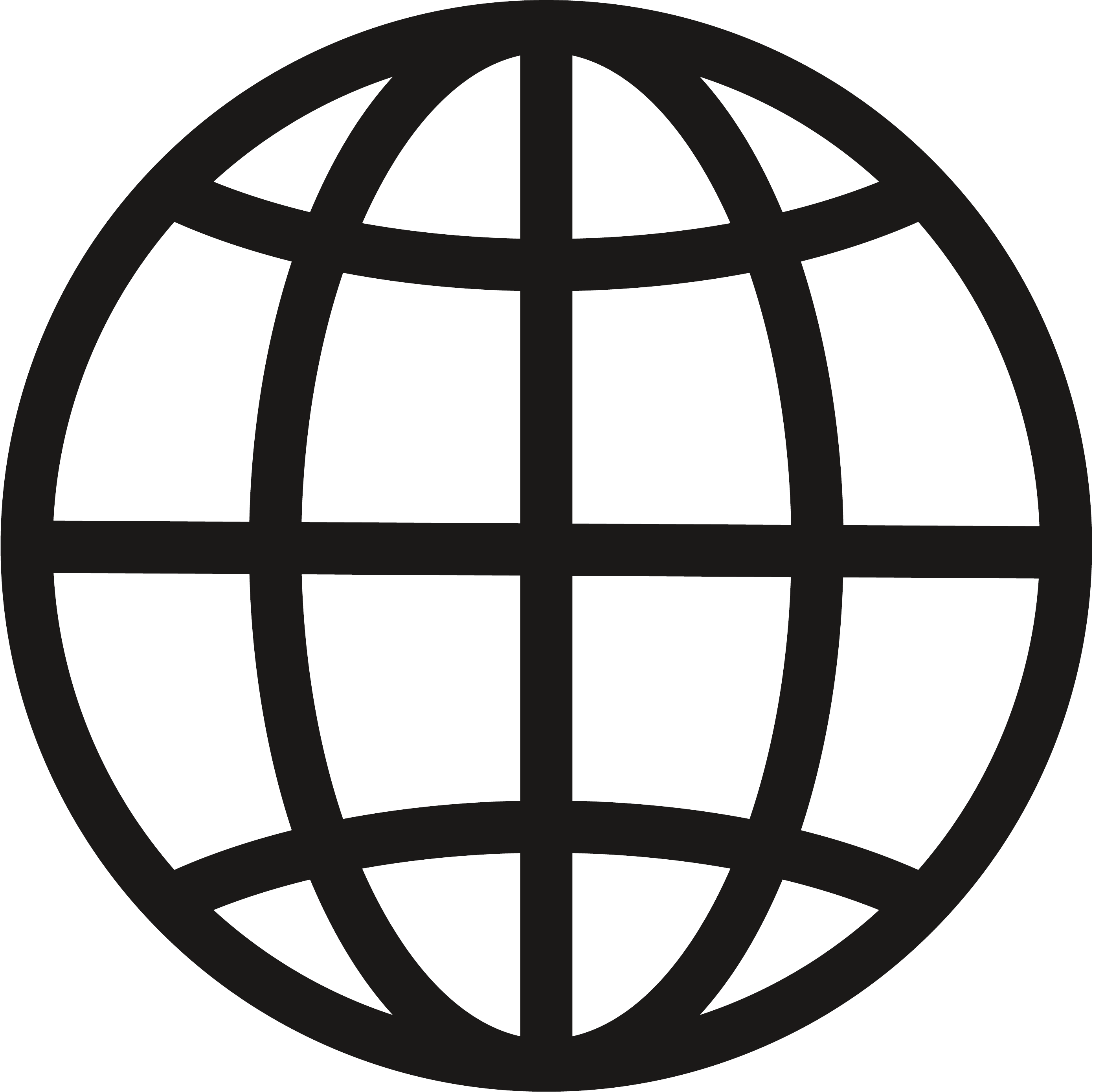}}\xspace}
\newcommand{\hf}{\raisebox{-1.3pt}{\includegraphics[height=1.05em]{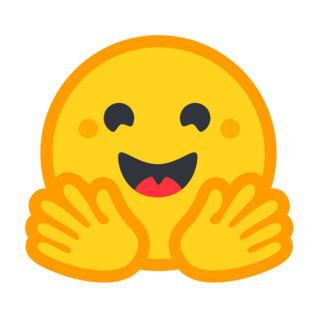}}\xspace}

\makeatletter
\fancypagestyle{firststyle}{
  \fancyhead[L]{}
  \fancyhead[C]{}
  \fancyhead[R]{}
  \fancyfoot[L]{\footerfont \the\correspondingauthor}
  \fancyfoot[R]{}
}
\makeatother

\DeclareTextFontCommand{\textbf}{\bfseries}

\title{\centering \fontsize{18}{24}\selectfont{VisionFoundry: Teaching VLMs Visual Perception \\ with Synthetic Images}}

\author{
    \vspace{.2cm}
    \parbox{\textwidth}{\centering
        Guanyu Zhou\textsuperscript{1} \quad
        Yida Yin\textsuperscript{1} \quad
        Wenhao Chai\textsuperscript{1} \quad
        Shengbang Tong\textsuperscript{2} \quad
        Xingyu Fu\textsuperscript{1} \quad
        Zhuang Liu\textsuperscript{1}
    }
    \\
    \vspace{.3cm}
    {\normalfont\fontsize{11}{15}\selectfont {\textsuperscript{1}Princeton University}\hspace{.1cm}}
    {\normalfont\fontsize{11}{15}\selectfont {\textsuperscript{2}New York University}}
    \\
    \vspace{.5cm}
    {\normalfont\fontsize{10}{13}\selectfont
      \href{https://zlab-princeton.github.io/VisionFoundry}{\worldwideweb~Project Page}
      \hspace{0.9cm}
      \href{https://github.com/zlab-princeton/VisionFoundry}{\github~Code}
      \hspace{0.9cm}
      \href{https://huggingface.co/datasets/zlab-princeton/VisionFoundry-10K}{\hf~Dataset}
    }
    \vspace{-.5cm}
}

\makeatletter
\renewcommand{\abscontent}{}%
\makeatother

\newenvironment{abstractblock}{%
  {\centering\large\bfseries Abstract\par}
  \vspace{0.2em}
  \begin{list}{}{%
      \setlength{\leftmargin}{2em}
      \setlength{\rightmargin}{2em}
      \setlength{\topsep}{0pt}
      \setlength{\parsep}{0pt}
  }
  \item[]
}{%
  \end{list}
  \par\normalfont\vspace{1em}
}

\begin{document}

\begingroup
\makeatletter
\let\raggedright\centering
\makeatother

\maketitle
\endgroup

\newcommand{\abstractcontent}{%
Vision-language models (VLMs) still struggle with visual perception tasks such as spatial understanding and viewpoint recognition, largely because natural image datasets provide limited supervision for low-level visual skills.
Can targeted synthetic supervision address these weaknesses without reference images or manual annotation?
To investigate this, we introduce \dataengine, an automated pipeline that takes only a task name as input, uses LLMs to synthesize paired questions, answers, and text-to-image (T2I) prompts, generates images with T2I models, and filters samples via multimodal verification.
With \dataengine, we construct \datasetname, a synthetic VQA dataset spanning 10 perception tasks.
Finetuning on \datasetname consistently improves perception benchmarks across three open-source backbones (e.g., +6.7\% on MMVP-pair and +10.5\% on CV-Bench-3D for Qwen2.5-VL-3B-Instruct) while preserving broader capabilities and showing positive data scaling.
The same synthetic supervision also yields consistent gains under reinforcement learning (RL) across all three backbones, and the framework remains effective under open-source synthesis and self-verification.
Our findings demonstrate that automated synthetic supervision offers an effective and scalable path toward systematic VLM training.
}

\newcommand{\linkstablestyleone}{%
    \begin{center}
        \small
        \renewcommand{\arraystretch}{1.2}
        \begin{tabular}{rll}
            \github & \textbf{Code} & \url{https://github.com/zlab-princeton/VisionFoundry}
        \end{tabular}
    \end{center}%
}

\newcommand{\linkstablestyletwo}{%
    \noindent\small
    \textbf{Code:} \url{https://github.com/zlab-princeton/VisionFoundry}
}

\begin{abstractblock}
\abstractcontent
\end{abstractblock}

\vspace{0.3cm}

\section{Introduction}
\label{sec:introduction}

Vision-language models (VLMs) have rapidly evolved into general-purpose systems capable of processing and reasoning over interleaved text and images~\citep{liu2023llava,bai2023qwenvl,dai2023instructblip,zhu2023minigpt4}.
They now support a broad spectrum of applications, from visual question answering \citep{liu2023llava,liu2023mmbench} and optical character recognition \citep{liu2023ocrbench} to multimodal reasoning \citep{yue2023mmmu,chen2024mmstar}, graphical user interface (GUI) grounding \citep{li2025screenspotpro}, and mathematical problem-solving \citep{lu2024mathvista}.
Despite these advances, VLMs still exhibit persistent weaknesses in visual perception, as highlighted by recent diagnostic benchmarks (e.g., MMVP, CV-Bench, and RealWorldQA). \citet{tong2024mmvp,tong2024cambrian} construct benchmarks that decouple visual perception from pure language priors, while RealWorldQA~\citep{xai2024realworldqa_dataset} stress-tests geometric and spatial reasoning.

One plausible contributing factor behind this perception bottleneck is limited supervision for these low-level visual skills in natural image datasets.
Natural image--text corpora, while vast, may not systematically cover the combinatorial range of variations required for robust visual perception.
This motivates a practical question: can we synthesize targeted supervision to address these weaknesses without relying on reference images or expensive human annotation?

Synthetic data is a core ingredient in LLM training. It can be generated on demand, targeted to specific failures, and iterated rapidly \citep{Gunasekar2023,Abdin2024,qinscaling}. A similar trend is emerging in vision, where text-to-image (T2I) models enable synthetic-caption generation and image--text corpus construction \citep{tian2023stablerep,fan2024scalinglaws,sharifzadeh2024synth2}. They also support multimodal tuning focused on specific visual skills \citep{liu2024synthvlm,jiao2024imgdiff,li2025sparcl,wu2025gift}. Synthetic data is near-unbounded and highly controllable, allowing systematic coverage of entity, attribute, relation, viewpoint, and style combinations that are rare in natural corpora and otherwise expensive to curate. For VLMs, the more useful question is what makes synthetic images good supervision: they should provide task-specific, reliable, and scalable training signals without reference images or expensive human annotation, while improving visual perception without noticeably degrading broader capabilities \citep{singh2024syntheticrobustness,hu2025modelcollapse,stan2025reasonfailures,wu2025compact}.

\begin{figure}[t]
    \centering
    \includegraphics[width=\linewidth]{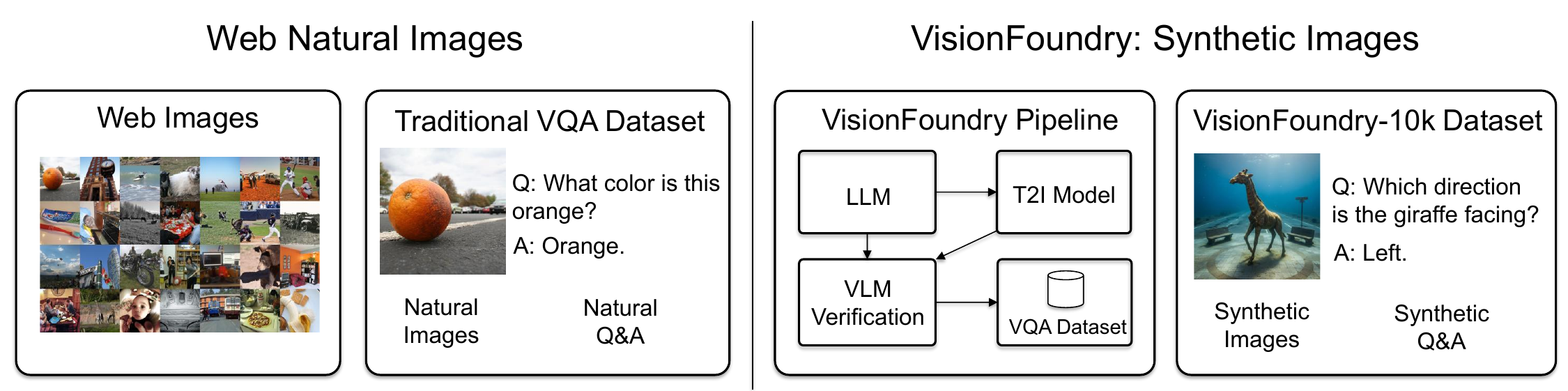}
    \caption{\textbf{Training VLMs with synthetic images.} \dataengine uses a task-keyword-only pipeline that requires no reference text or images: it generates task-aware T2I prompts and question--answer pairs, synthesizes images, and uses the resulting supervision to improve capabilities such as visual perception in VLMs. Compared with traditional web-image collection pipelines, this approach provides more controllable and task-targeted supervision.}
    \label{fig:main}
\end{figure}

\begin{figure*}[t]
    \centering
    \includegraphics[width=0.99\linewidth]{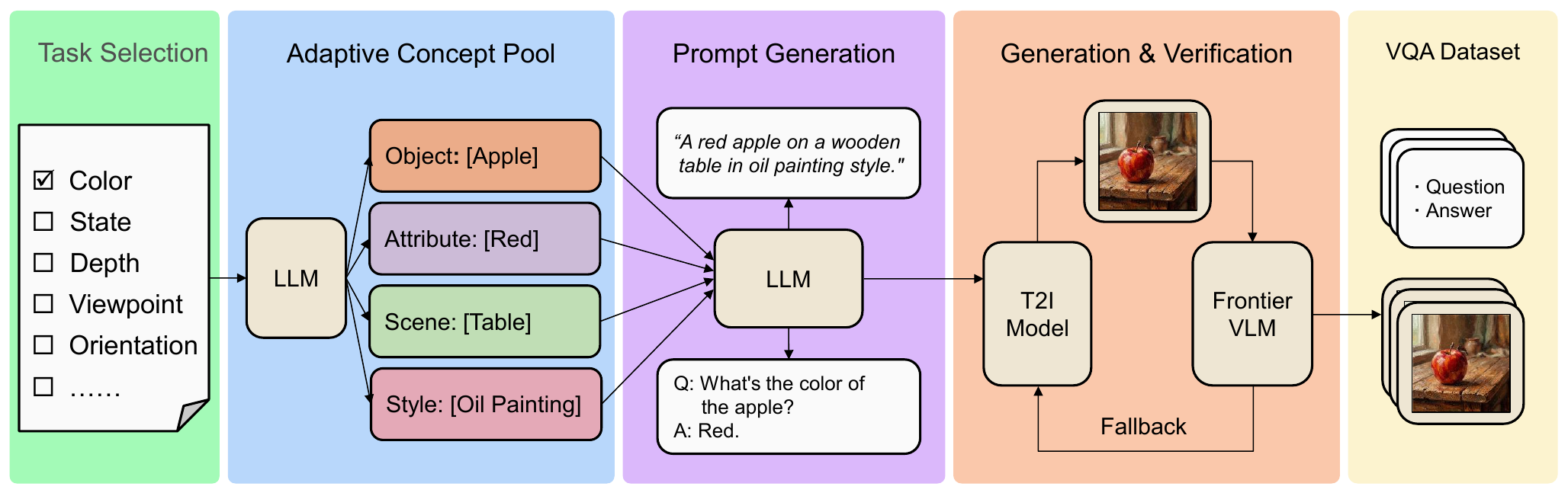}
    \caption{\textbf{\dataengine overview.} Given only task keywords, an LLM builds an adaptive concept pool; compositional sampling forms entities for T2I prompts; and a T2I model synthesizes images that a frontier VLM verifies to yield a high-quality VQA dataset.}
    \label{fig:pipeline}
\end{figure*}

We build \dataengine, an end-to-end synthetic data-generation pipeline that takes only a task name and produces verifier-filtered image--question--answer triples (see Figure~\ref{fig:main}). \dataengine composes three components into a closed-loop system (Figure~\ref{fig:pipeline}):
(1) a large language model generates question--answer pairs and detailed T2I prompts conditioned on the target task;
(2) a modern T2I model synthesizes images conditioned on those prompts; and
(3) a strong multimodal judge verifies alignment between the generated image and the answer-determining visual statement, filtering out misaligned samples.
This pipeline eliminates the need for reference images and manual labeling, reducing cost while using the native cross-modal capabilities of T2I models.

Using \dataengine, we construct \datasetname, a synthetic visual question answering (VQA) dataset of 10k image--question--answer triples spanning 10 low-level visual perception tasks (1k samples per task), including spatial understanding, relative depth, and viewpoint variation (Section~\ref{sec:dataset}). We use this controlled setting as a validation to isolate the direct contribution of synthetic images to visual capability learning, rather than gains primarily driven by text-heavy reasoning supervision.
These tasks are motivated by benchmark analyses that identify spatial and perceptual failures as persistent weaknesses of contemporary VLMs \citep{tong2024mmvp,tong2024cambrian,xai2024realworldqa_dataset}.

Across three representative open-source models, Qwen2.5-VL-3B-Instruct~\citep{bai2025qwen25vl}, Llama-3.2-11B-Vision-Instruct~\citep{meta2024llama32vision}, and MiMo-VL-7B-SFT~\citep{coreteam2025mimovltechnicalreport}, finetuning on \datasetname consistently improves visual perception benchmarks in our experiments, especially on Qwen2.5-VL-3B-Instruct, with gains such as +6.7\% on MMVP-pair and +10.5\% on CV-Bench-3D, while preserving broader capabilities and showing an upward trend as the data size increases (Section~\ref{sec:experiments}).
We also evaluate on broader multimodal reasoning, OCR, and GUI grounding benchmarks, where models maintain competitive general capabilities.
We further observe that an equal-sized synthetic--natural mixture outperforms pure natural subsets, suggesting that synthetic supervision provides complementary signals that are hard to obtain from limited real data alone \citep{tian2023stablerep,fan2024scalinglaws,singh2024syntheticrobustness}.
Beyond supervised finetuning (SFT), we further find that the same synthetic supervision improves visual perception under RL.

Our results suggest that limited task-targeted supervision is an important factor behind the perception bottleneck in VLMs and that synthetic data curation can meaningfully alleviate part of this weakness.
Beyond immediate benchmark improvements, \dataengine represents a systematic step toward more systematic synthetic-data workflows for training VLMs.
Our results further suggest that synthetic images can serve as a useful primitive for post-training beyond SFT, addressing capability gaps as new evaluations emerge. They may also support pretraining regimes with broader compositional coverage than web-scale natural images can provide.
We hope this work can catalyze a broader shift toward fully automated synthetic multimodal data as a promising path to stronger VLMs.

\section{\dataengine}
\label{sec:data_engine}

\dataengine is a task-aware synthetic data generation pipeline that produces high-quality VQA supervision for VLMs using only high-level task specifications, without reference images, manual QA annotations, or real image--caption pairs.
The methodology is guided by three core principles: \emph{controllability}, \emph{visual determinism}, and \emph{verification}.
\emph{Controllability} leverages explicit task configurations and entity pools to systematically cover targeted capabilities.
\emph{Visual determinism} encodes answer-determining visual facts directly into the prompt so that questions are answerable solely from the synthesized image.
\emph{Verification} employs an automated verifier to filter subtle misalignments between text specifications and visual content that would otherwise inject noise into supervision.

Operationally, given a high-level task specification (e.g., spatial understanding, counting, depth reasoning), \dataengine has three stages:
(1) task-aware generation of questions, answers, and T2I prompts;
(2) image synthesis conditioned on the generated prompts; and
(3) alignment verification and filtering using a strong multimodal judge.
Together, these stages produce supervision that is visually grounded, unambiguous, and robust. Appendix~\ref{app:opensource_t2i} and Appendix~\ref{app:opensource_verifier} show that \dataengine remains effective with open-source alternatives.
Figure~\ref{fig:pipeline} provides an overview of the pipeline, with its left, middle, and right parts corresponding to Sections~\ref{subsec:qa_prompt}, \ref{subsec:image_synthesis}, and~\ref{subsec:alignment}, respectively.
We provide the exact pipeline prompts in the appendix for transparency and reproducibility in downstream audits.

\subsection{VQA Triplet Generation}
\label{subsec:qa_prompt}

\begin{figure*}[t]
    \centering
    \includegraphics[width=0.99\linewidth]{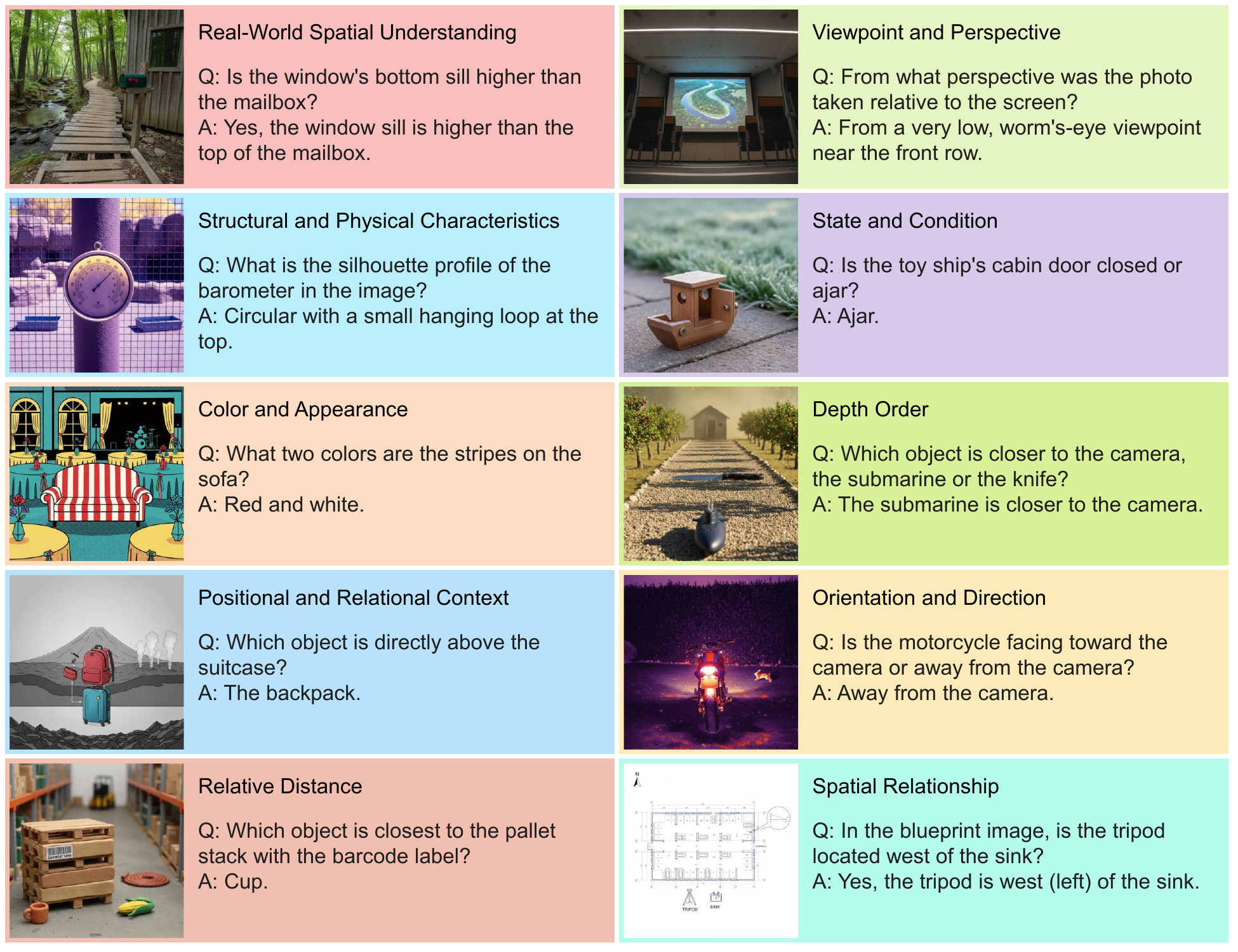}
    \caption{\textbf{\datasetnamenosym examples.} Randomly selected examples from our dataset, covering all 10 tasks. The task names are annotated at the top and serve as the only valid input to the pipeline. Each panel shows a generated image, its corresponding question, and a ground-truth answer.}
    \label{fig:dataset_examples}
\end{figure*}

\paragraph{Task-aware generation.}
The left part of Figure~\ref{fig:pipeline} corresponds to the first stage of \dataengine, which constructs a triple consisting of a question, its answer, and a corresponding T2I prompt. This stage is driven by GPT-5.2~\citep{openai2025gpt52} and conditions generation on an explicit \emph{task configuration}.
Each task specifies the target capability (e.g., spatial relations, color discrimination), the number of objects involved, and optional constraints such as required attributes or relations.
To support systematic coverage, we construct an \emph{entity pool} composed of objects, attributes, scenes, styles, and task-specific custom dimensions.
Entities are randomly sampled from the Cartesian product of these dimensions, yielding structured visual configurations.

Given a sampled entity, the LLM is prompted to generate:
(i) a question whose answer is fully determined by visible content,
(ii) a concise and deterministic answer, and
(iii) a detailed T2I prompt that explicitly encodes the answer-determining visual facts.
To avoid ambiguity, the LLM is instructed to rely exclusively on visually verifiable properties and to avoid commonsense assumptions.

\paragraph{Design rationale.}
This design enforces tight coupling between language supervision and visual content at image generation time.
By explicitly embedding the correct answer directly into the T2I prompt, we reduce the risk of producing images that are irrelevant or underspecified for the question. This is a common failure mode in synthetic multimodal data.

\subsection{Image Synthesis}
\label{subsec:image_synthesis}

\paragraph{Prompt-conditioned generation.}
The second stage of \dataengine synthesizes images conditioned on the T2I prompts (Figure~\ref{fig:pipeline}, middle).
We use Gemini-2.5-Flash-Image~\citep{google2025gemini25flashimage,comanici2025gemini25} as the generator.
Each image is generated from its corresponding prompt, which specifies key objects, attributes, spatial arrangements, and visual styles to ensure that the image contains all information required to answer the associated question.
Our pipeline does not rely on model-specific internals, ensuring straightforward compatibility with alternative generators.

\paragraph{Selective iterative refinement.}
To improve yield, \dataengine allows limited iterative refinement of images.
If an initially generated image fails downstream verification, the system can request a localized edit that minimally modifies the image to better satisfy the intended visual statement.
This refinement process is conceptually similar to prompt-based image editing, but is only used when necessary and does not alter the original question and answer pair.

\begin{table*}[t]
\centering
\resizebox{\textwidth}{!}{
{\footnotesize
\begin{tabular}{l>{\centering\arraybackslash}m{1.42cm}>{\columncolor{\synthcol}\centering\arraybackslash}m{1.42cm}>{\centering\arraybackslash}m{1.42cm}>{\columncolor{\synthcol}\centering\arraybackslash}m{1.42cm}>{\centering\arraybackslash}m{1.42cm}>{\columncolor{\synthcol}\centering\arraybackslash}m{1.42cm}}
\toprule
\multirow{2}{*}{\textbf{Benchmark}} & 
\multicolumn{2}{c}{\footnotesize\textbf{\shortstack{Qwen2.5-VL-3B\\Instruct}}} & 
\multicolumn{2}{c}{\footnotesize\textbf{\shortstack{MiMo-VL-7B\\SFT}}} & 
\multicolumn{2}{c}{\footnotesize\textbf{\shortstack{Llama-3.2-11B\\Vision-Instruct}}} \\
\cmidrule(lr){2-3} \cmidrule(lr){4-5} \cmidrule(lr){6-7}
& Baseline & Synth & Baseline & Synth & Baseline & Synth \\
\midrule
\multicolumn{7}{l}{\textbf{Visual Perception Benchmarks}} \\
\midrule
$\mathrm{MMVP}_{\scriptstyle\text{pair}}$~\citep{tong2024mmvp} & 35.3 & \textbf{42.0} & 43.3 & \textbf{57.3} & 42.7 & \textbf{46.7} \\
$\mathrm{MMVP}_{\scriptstyle\text{single}}$~\citep{tong2024mmvp} & 64.3 & \textbf{68.3} & 66.7 & \textbf{77.7} & 70.3 & \textbf{71.7} \\
CV-Bench-2D~\citep{tong2024cambrian} & 67.3 & \textbf{72.4} & 74.3 & \textbf{79.0} & 70.4 & \textbf{71.7} \\
CV-Bench-3D~\citep{tong2024cambrian} & 66.0 & \textbf{76.5} & 72.3 & \textbf{83.7} & 74.4 & \textbf{75.3} \\
RealWorldQA~\citep{xai2024realworldqa_dataset} & 65.0 & \textbf{66.9} & 65.9 & \textbf{67.5} & 63.0 & \textbf{64.6} \\
\midrule
\multicolumn{7}{l}{\textbf{General Understanding and Reasoning Benchmarks}} \\
\midrule
BLINK~\citep{fu2024blink} & 48.1 & 47.9 & 58.9 & 58.7 & 34.3 & \textbf{35.5} \\
$\mathrm{MMMU}_{\scriptstyle\text{Val}}$~\citep{yue2023mmmu} & 48.0 & 47.1 & 49.4 & \textbf{52.3} & 40.1 & \textbf{40.7} \\
$\mathrm{MMBench}_{\scriptstyle\text{EN\ v1.1}}$~\citep{liu2023mmbench} & 75.8 & \textbf{76.1} & 50.5 & \textbf{81.6} & 66.6 & \textbf{73.8} \\
MMStar~\citep{chen2024mmstar} & 55.7 & 54.5 & 53.7 & \textbf{62.7} & 49.0 & \textbf{50.9} \\
$\mathrm{MathVista}_{\scriptstyle\text{Mini}}$~\citep{lu2024mathvista} & 62.3 & \textbf{62.9} & 39.1 & \textbf{52.4} & 47.0 & \textbf{48.3} \\
\midrule
\multicolumn{7}{l}{\textbf{Application-Specific Benchmarks}} \\
\midrule
$\mathrm{ScreenSpot\text{-}Pro}_{\scriptstyle\text{avg}}$~\citep{li2025screenspotpro} & 19.4 & \textbf{20.2} & 3.4 & \textbf{5.8} & 0.0 & \textbf{6.4} \\
OCRBench~\citep{liu2023ocrbench} & 82.7 & 82.4 & 83.3 & 82.9 & 75.4 & 75.2 \\
MMSI-Bench~\citep{yang2025mmsibench} & 9.2 & \textbf{9.9} & 8.4 & \textbf{11.5} & 3.9 & \textbf{8.9} \\
3DSRBench~\citep{ma2025_3dsrbench} & 35.2 & \textbf{36.8} & 32.0 & \textbf{40.0} & 27.5 & \textbf{39.2} \\
LEGO~\citep{tang2025legopuzzles} & 15.8 & 14.4 & 24.0 & \textbf{28.8} & 5.3 & \textbf{6.2} \\
\bottomrule
\end{tabular}
}
}
\vspace{10pt}
\caption{\textbf{Main Benchmark Results.} Main results on 13 benchmarks with 15 reported metrics across three VLMs. Models trained on \datasetname improve visual perception performance while maintaining general capabilities. We report the mean score over four independent runs. Visual perception benchmarks are highlighted at the top.}
\label{tab:main_results_13bench}
\end{table*}

\begin{table}[t]
\centering
\small
\resizebox{\linewidth}{!}{%
\begin{tabular}{lccccc}
\toprule
\textbf{Model} & $\mathbf{MMVP}_{\scriptstyle\text{pair}}$ & $\mathbf{MMVP}_{\scriptstyle\text{single}}$ & \textbf{CV-Bench-2D} & \textbf{CV-Bench-3D} & \textbf{RealWorldQA} \\
\midrule
Qwen2.5-VL-3B-Instruct & 46.0 (+10.7) & 71.3 (+7.0) & 72.9 (+5.6) & 72.3 (+6.3) & 67.2 (+2.2) \\
Llama-3.2-11B-Vision-Instruct & 45.3 (+2.6) & 70.7 (+0.4) & 72.4 (+2.0) & 76.3 (+1.9) & 63.8 (+0.8) \\
MiMo-VL-7B-SFT & 48.0 (+4.7) & 73.3 (+6.6) & 78.7 (+4.4) & 75.8 (+3.5) & 69.2 (+3.3) \\
\bottomrule
\end{tabular}%
}
\vspace{10pt}
\caption{\textbf{RL results on visual perception benchmarks.} Evaluation of GRPO across three VLM backbones on the 10-task \datasetname dataset. Numbers in parentheses indicate absolute gains over each model's baseline.}
\label{tab:rl_results_perception}
\end{table}

\subsection{Alignment Verification and Filtering}
\label{subsec:alignment}

\paragraph{Statement-based verification.}
In the final stage of \dataengine (Figure~\ref{fig:pipeline}, right), the pipeline verifies whether each synthesized image is strictly consistent with its corresponding question and answer, ensuring dataset quality while remaining fully automatic.
Given a question and candidate answer, we convert them into a short declarative visual statement capturing the key visual fact (e.g., ``The red cube is to the left of the blue sphere'').
A strong multimodal judge, instantiated as Gemini-3-Pro~\citep{google2025gemini3pro}, inspects the image and statement, returning a binary accept/reject decision based on whether the core visual relation is present while ignoring minor stylistic variations.
Importantly, this verification stage is fully automated and involves no manual intervention.

\paragraph{Binary filtering criterion.}
Only samples for which the judge confirms alignment are retained.
If an image fails verification after limited refinement attempts, the entire sample is discarded and a new entity is sampled.
By removing human annotation and manual correction, our design deliberately prioritizes end-to-end automation and scalability, allowing us to directly evaluate the viability and scaling dynamics of fully automated synthetic multimodal supervision.

\paragraph{Why verification matters.}
Without explicit verification, synthetic datasets are prone to subtle misalignments, such as missing objects, incorrect spatial relations, or visually ambiguous scenes.
By incorporating a multimodal verifier, \dataengine provides an automated filtering stage that screens out inconsistent samples before downstream training.
Empirically, current multimodal models provide reliable alignment signals for these visual distinctions (detailed verification-accuracy records are provided in Appendix~\ref{app:data_quality}).
Appendix~\ref{app:opensource_verifier} further shows that the downstream Qwen backbone itself can serve as the verifier: a full 10-task dataset constructed with Qwen-based image synthesis and self-verification still improves all five visual-perception benchmarks after finetuning.

The output of the pipeline is a fully synthetic collection of image--question--answer triples, formatted for instruction tuning.
Each retained sample is verifier-approved and intended to be visually grounded and answerable solely from the image.
Overall, \dataengine demonstrates how modern LLMs and T2I models can be composed into a principled system for synthetic multimodal data generation and provides a practical framework for studying improvements in current VLMs' visual understanding.

\section{\datasetnamenosym Dataset}
\label{sec:dataset}

Using \dataengine, we construct \datasetname, a synthetic VQA dataset targeting visual perception.
\datasetname contains 10k image--question--answer triples across 10 tasks, with 1k cases per task.
All images are generated by T2I models and filtered through automated multimodal alignment verification as described in Section~\ref{sec:data_engine}.
Figure~\ref{fig:dataset_examples} shows \datasetname examples.

\paragraph{Task selection.}
The tasks are chosen to emphasize visual perception skills that are known to be challenging for current VLMs and are weakly correlated with pure language ability.
We treat \datasetname as a first-step validation of synthetic images as training supervision. We therefore prioritize visual perception tasks over text-heavy or long-chain reasoning tasks to isolate the visual learning effect from improvements that come mainly from textual reasoning signals.
In particular, we focus on perception-oriented tasks that depend on spatial layout, depth, and relative attributes, rather than high-level semantics or commonsense knowledge.
The 10 tasks included in \datasetname are \emph{Orientation and Direction}, \emph{Viewpoint and Perspective}, \emph{Positional and Relational Context}, \emph{Spatial Relationship}, \emph{State and Condition}, \emph{Structural and Physical Characteristics}, \emph{Color and Appearance}, \emph{Depth Order}, \emph{Relative Distance}, and \emph{Real-World Spatial Understanding}.

\begin{figure*}[t]
    \centering
    \includegraphics[width=0.99\linewidth]{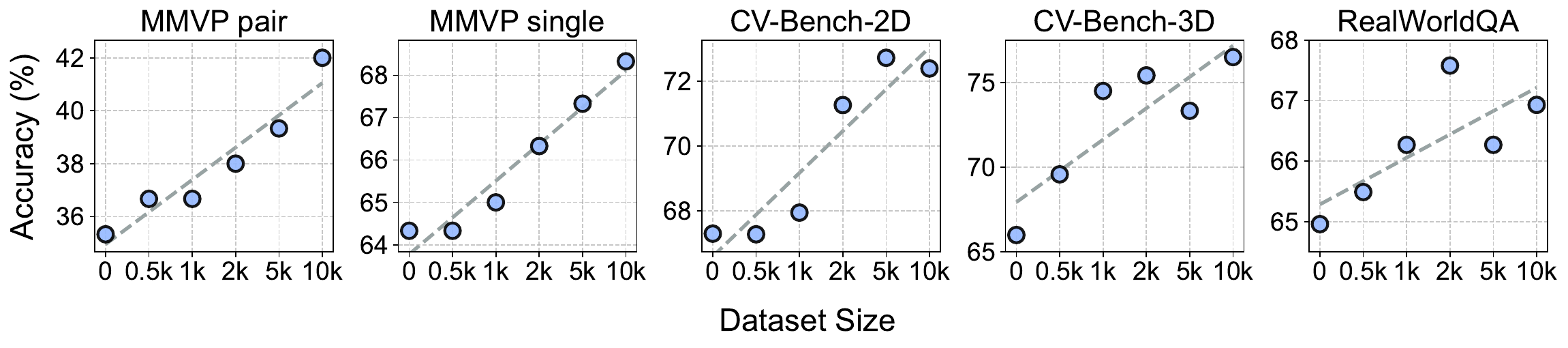}
    \caption{\textbf{Data-Size Effects of Synthetic Supervision.} We report performance trends as the amount of \dataengine\ synthetic VQA supervision increases.
    ``10k'' corresponds to the full \datasetname dataset (10k samples), while ``500/1k/2k/5k'' are random subsets sampled from it.
    The training recipe matches the main experimental setting.
    Overall, results show an upward trend on visual perception benchmarks as synthetic data increases.}
    \label{fig:scaling}
\end{figure*}

\paragraph{Dataset construction and statistics.}
For each task, we define a task-specific configuration that constrains the number of objects, admissible attributes, and relational predicates.
Entity configurations are systematically sampled from an underlying entity pool, ensuring diversity in object categories, attributes, scenes, and visual styles.
For every sampled configuration, \dataengine generates a single question with a deterministic answer, plus a corresponding T2I prompt encoding the answer-determining visual facts.
After image synthesis and alignment verification, exactly 1k verifier-accepted samples are retained per task, resulting in the balanced \datasetname dataset of 10k examples.
All questions are short and unambiguous, and all answers are concise (e.g., binary, categorical, or short phrases), enabling reliable finetuning and evaluation.

\paragraph{Positioning vs.\ existing datasets.}
Unlike natural-image VQA datasets, \datasetname is not designed to maximize semantic coverage or linguistic diversity.
Instead, it prioritizes precise visual grounding and controlled difficulty, making it suitable for diagnosing and improving core perceptual weaknesses of VLMs.
Compared with manually curated datasets, \datasetname is fully synthetic and automatically quality-controlled by verification.
As shown in our experiments, training on \datasetname yields consistent improvements on multiple visual perception benchmarks. This validates the effectiveness of targeted synthetic supervision for visual perception.

\section{Experiments}
\label{sec:experiments}

We evaluate the effectiveness of \dataengine-generated data for improving VLMs' visual perception.
Our experiments answer five questions:
(i) whether synthetic data generated by \dataengine improves performance on visual perception benchmarks;
(ii) whether such improvements generalize across different VLM backbones and model scales;
(iii) whether gains can be achieved without noticeably affecting performance on general-purpose tasks;
(iv) how performance changes with data scale and epochs; and
(v) whether the same synthetic data remains effective under RL.

\subsection{Experimental Setup}
\label{subsec:exp_setup}

\paragraph{Models.}
We conduct experiments on three representative open-source VLMs covering different scales:
Qwen2.5-VL-3B-Instruct~\citep{bai2025qwen25vl},
Llama-3.2-11B-Vision-Instruct~\citep{meta2024llama32vision,dubey2024llama3},
and MiMo-VL-7B-SFT~\citep{coreteam2025mimovltechnicalreport}.
For the main experiments, all models are finetuned without LoRA; model-specific settings and hyperparameters are given in Appendix~\ref{app:opt_details}.
We additionally evaluate RL post-training on the same data, with settings provided in Appendix~\ref{app:rl_details}.

\paragraph{Training data.}
The \datasetname dataset consists of 10k image--question--answer triples generated by \dataengine, evenly distributed across 10 visual perception tasks (Section~\ref{sec:dataset}).
For the main experiments, we train on \datasetname alone for one epoch.

\begin{figure*}[t]
    \centering
    \includegraphics[width=0.97\linewidth]{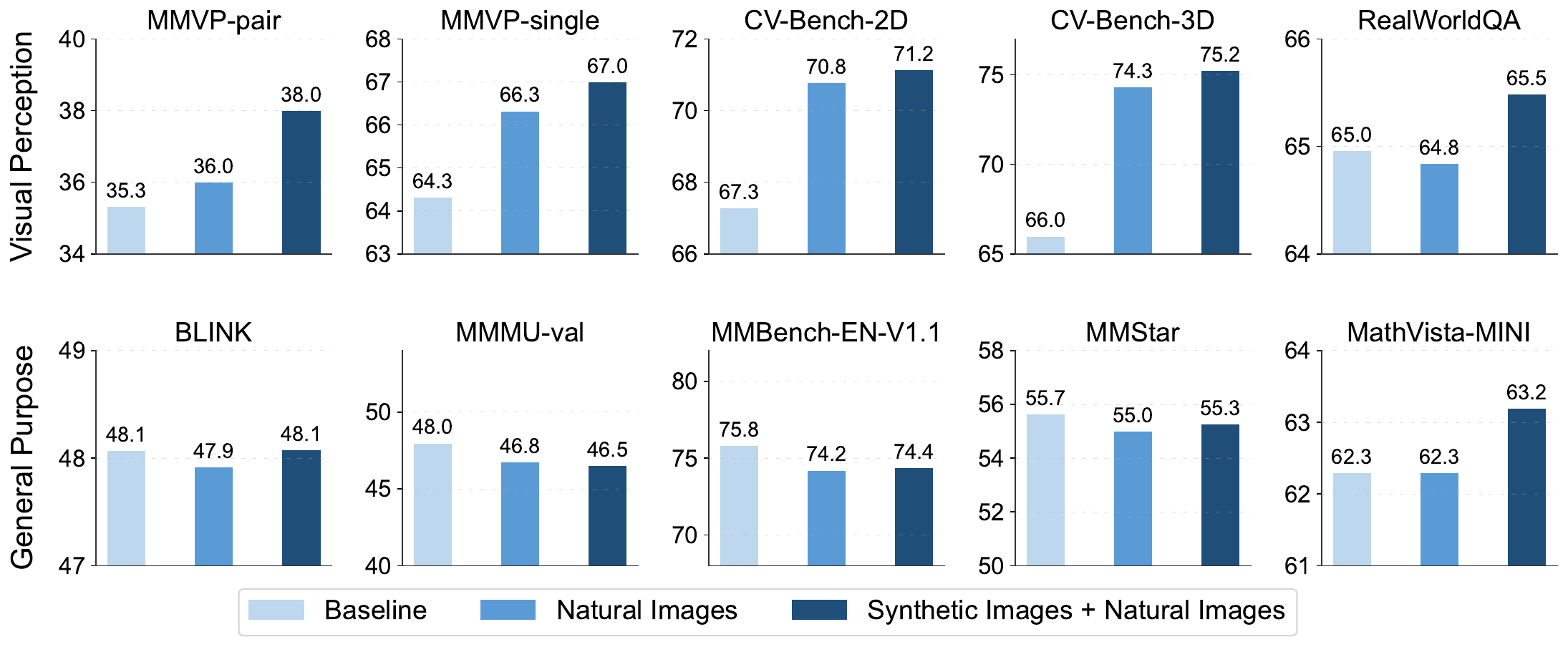}
    \vspace{10pt}
    \caption{\textbf{Equal-sized mixture vs.\ pure natural data.} 
    We compare training on an equal-sized mix of \datasetname samples and natural data against training on a size-matched natural-only subset.
    In the figure, the bold-bordered first row denotes visual perception benchmarks, while the second row reports general-purpose benchmarks.
    The mixed setting yields consistently higher accuracy on visual perception benchmarks while maintaining comparable performance on general-purpose evaluations.}
    \label{fig:mixed_vs_natural}
\end{figure*}
\vspace{-5pt}

\paragraph{Benchmarks and evaluation protocol.}
We evaluate models on 13 widely used benchmarks spanning three categories: visual perception, general reasoning, and application-specific tasks, using the VLMEvalKit framework~\citep{duan2024vlmevalkit}.
MMSI-Bench and LEGO use VLMEvalKit's default circular setting.
For 3DSRBench, we report VLMEvalKit's default FlipEval result in circular-evaluation mode.
Since our primary focus is visual perception, we evaluate MiMo-VL-7B-SFT in non-thinking mode (detailed settings in Appendix~\ref{app:opt_details}).
All reported numbers for the main SFT experiments in Table~\ref{tab:main_results_13bench} are averaged over four independent runs.

\subsection{Main Results}
\label{subsec:main_results}

Table~\ref{tab:main_results_13bench} summarizes the main results across all 13 benchmarks.
For all VLMs, training with \dataengine synthetic data consistently improves performance on visual perception benchmarks, while general-purpose benchmarks exhibit more varied, model-dependent responses across tasks.

\paragraph{Visual perception gains.}
The largest improvements are observed on MMVP, CV-Bench, and RealWorldQA, which explicitly stress spatial reasoning, depth ordering, attribute recognition, and other low-level visual perception skills.
These gains validate that \dataengine effectively targets core visual weaknesses identified in prior work~\citep{tong2024mmvp,tong2024cambrian}.

\vspace{10pt}
\paragraph{General-purpose tasks.}
On general-purpose and application-specific benchmarks, outcomes vary across models and benchmarks; for example, Llama improves on 3DSRBench, while MiMo-VL-7B-SFT exhibits particularly pronounced gains on MMBench and MathVista, suggesting that enhanced perceptual grounding provides substantial support on multimodal benchmarks for this backbone. On OCRBench, performance remains largely stable with minor variations, reflecting the domain boundary between low-level visual perception and document text recognition. While perceptual benchmarks show uniform gains, general-purpose metrics exhibit more variation across models, indicating that targeted perception tuning does not compromise overall multimodal reasoning.

\paragraph{Reinforcement learning.}
We next test whether the utility of \dataengine supervision is specific to SFT. Using the same dataset, we apply GRPO to all three backbones. As shown in Table~\ref{tab:rl_results_perception}, RL improves performance on all visual perception benchmarks. The gains are particularly large on MMVP and CV-Bench, indicating that the targeted synthetic supervision remains useful under a different post-training objective.

\subsection{Data-Size Sensitivity}
\label{subsec:scaling_laws}

We analyze how model performance changes with the amount of synthetic data.
Instead of varying task types, we vary only the number of synthetic image--question--answer triples.
Concretely, we train on subsets of size 0.5k, 1k, 2k, and 5k, as well as the full 10k dataset.
All subsets are obtained by random sampling. This lets us compare performance across a consistent sequence of subset sizes, while remaining agnostic about how the trend might vary under other random draws.

\begin{figure*}[t]
    \centering
    \includegraphics[width=1.0\linewidth]{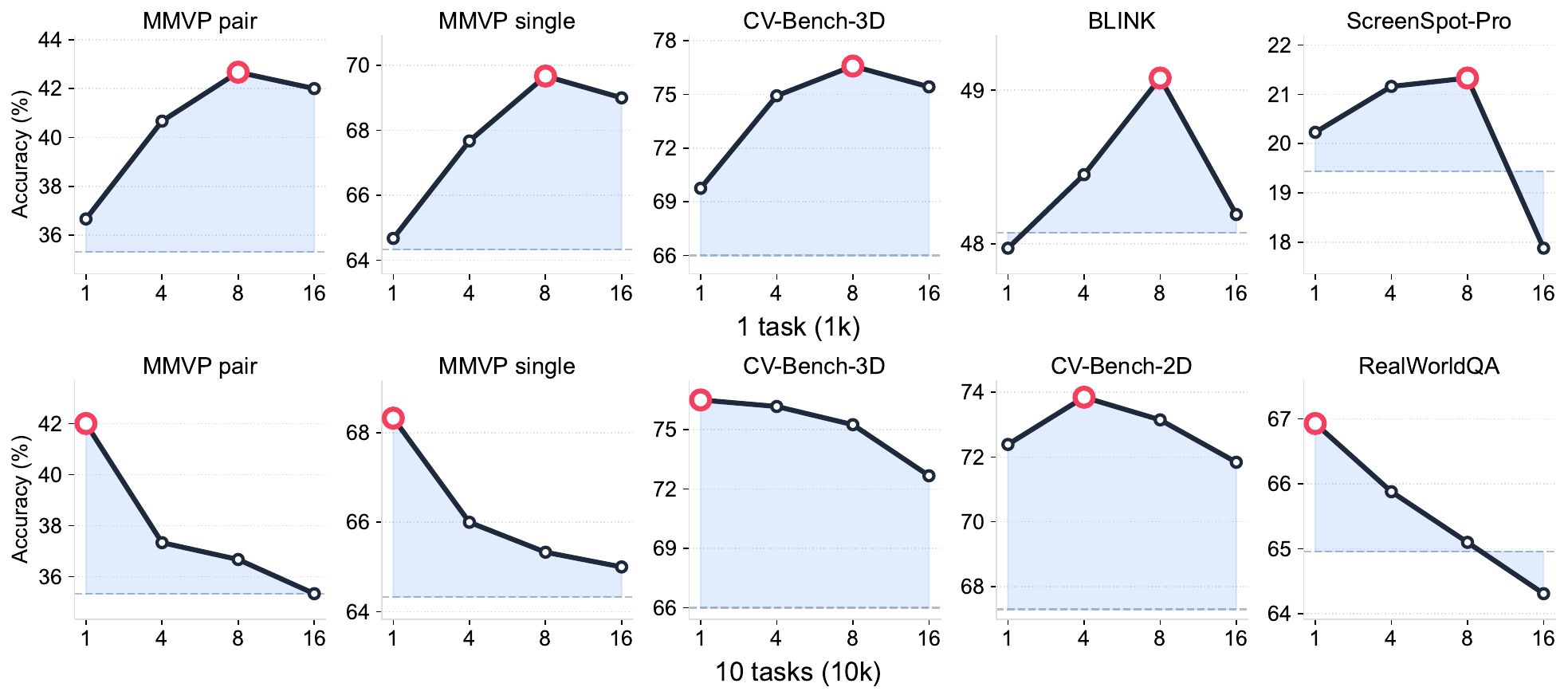}
    \caption{\textbf{Epoch trade-off.} The first row uses 1k samples from one randomly selected task, while the second row uses the full \datasetname. With only a single-task 1k subset, performance typically converges after around 8 epochs. With the full 10-task set, convergence is reached with fewer training epochs. The dashed line indicates the baseline.}
    \vspace{-10pt}
    \label{fig:epochs}
\end{figure*}

\paragraph{Setup.}
All experiments use Qwen2.5-VL-3B-Instruct. This model offers headroom and reveals data-size effects clearly in the low-data regime. We follow the training setup from the main experiments. We use synthetic-only finetuning under identical hyperparameters described in Appendix~\ref{app:opt_details}.

\paragraph{Results and conclusions.}
Figure~\ref{fig:scaling} shows consistent gains on visual perception benchmarks as the synthetic data budget increases.
At the full 10k setting, we observe sizeable gains over the baseline on key diagnostics: MMVP-pair improves from 35.3 to 42.0 (+6.7), MMVP-single from 64.3 to 68.3 (+4.0), CV-Bench-2D from 67.3 to 72.4 (+5.1), and CV-Bench-3D from 66.0 to 76.5 (+10.5).
RealWorldQA also improves from 65.0 to 66.9 (+1.9).
A few curves exhibit mild non-monotonicity (e.g., due to subset sampling variance). However, the overall trend is robust: more verifier-filtered synthetic supervision consistently translates into better low-level perception performance.
The largest gains appear on 3D/spatial diagnostics (e.g., CV-Bench-3D), suggesting that \dataengine\ provides especially effective coverage for geometry- and relation-heavy visual skills.

\subsection{Further Analysis}
\label{subsec:further_analysis}

\paragraph{Mix vs.\ natural.}
\label{subsubsec:mix_vs_natural}
We compare training on an equal-sized mixture of synthetic and natural data against training on pure natural data.
Specifically, we construct:
(i) a mixed dataset of 2k \datasetname samples and 2k randomly sampled LLaVA-Instruct-80K samples (4k in total); and
(ii) a pure natural dataset of 4k LLaVA-Instruct-80K samples, where the first 2k are the same natural samples used in (i) and kept in the same order, and the remaining 2k are additionally randomly sampled.
All experiments use Qwen2.5-VL-3B-Instruct and are trained for one epoch; full hyperparameters are provided in Appendix~\ref{app:opt_details}.

Results (Figure~\ref{fig:mixed_vs_natural}) show that the mixed dataset consistently outperforms pure natural data on visual perception benchmarks, while achieving comparable performance on other benchmarks.
This suggests that synthetic data provides complementary visual supervision that is difficult to obtain from small natural subsets alone and supports broader synthetic--natural training paradigms.

\paragraph{Training epochs.}
\label{subsubsec:epochs}
We analyze the effect of training epochs under two settings.
In the first setting, we use 1k samples from one randomly selected task; in the second setting, we use the full \datasetname (10 tasks).
For both settings, we train Qwen2.5-VL-3B-Instruct for 1, 4, 8, and 16 epochs.

Results (Figure~\ref{fig:epochs}) show that on the single-task 1k subset, performance improves up to around 8 epochs and saturates thereafter, whereas the full 10-task dataset reaches optimal performance with fewer epochs.
These observations provide practical guidance for choosing training schedules under different task-scale settings.

\begin{figure}[t]
    \centering
    \includegraphics[width=1.0\linewidth]{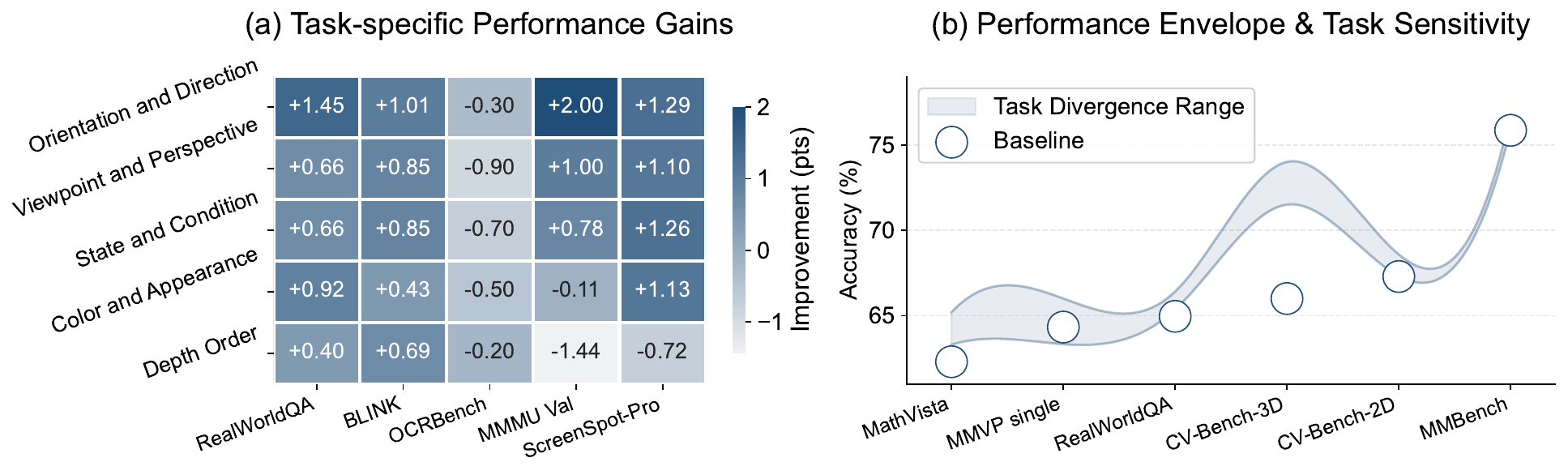}
    \vspace{5pt}
    \caption{\textbf{Task-wise gains and cross-benchmark divergence.} \textbf{(a):} Visualization of five representative task-specific 1k subsets; complete results for all ten task-specific models are reported in Appendix Tables~\ref{tab:taskwise_complete_part1} and~\ref{tab:taskwise_complete_part2}. Darker colors indicate larger gains, while lighter colors indicate smaller gains or regressions. \textbf{(b):} Per-benchmark response divergence across tasks, where a larger top-to-bottom accuracy span indicates stronger task sensitivity. While visual perception benchmarks generally benefit from synthetic supervision, the magnitude and direction of gains remain benchmark-dependent.}
    \label{fig:task1}
\end{figure}

\paragraph{Task-wise analysis.}
\label{subsubsec:task_boundaries}
To understand how different synthetic capabilities transfer across benchmarks, we perform a task-wise dissection of \datasetname and summarize representative patterns in Figure~\ref{fig:task1}.
Following the main training recipe, we train ten task-specific models, each on a 1k subset corresponding to one task, and evaluate all models on the same benchmark suite. Figure~\ref{fig:task1}(a) visualizes five representative tasks for readability, while complete results for all ten task-specific models are reported in Appendix Tables~\ref{tab:taskwise_complete_part1} and~\ref{tab:taskwise_complete_part2}.
Detailed experimental settings (identical to the main experiment except task-specific 1k subsets) are provided in Appendix~\ref{app:supp_protocols}.

The left panel of Figure~\ref{fig:task1} shows two clear patterns.
First, most visual perception benchmarks exhibit consistent improvements across tasks, supporting the effectiveness of task-aware synthetic supervision for visual understanding.
Second, OCRBench performance remains essentially stable across tasks, delineating the scope of perception-oriented data from document text understanding.

Importantly, benchmark responses are not uniform across tasks.
For example, \emph{Depth Order} provides strong gains on spatially grounded benchmarks. However, it yields limited transfer on ScreenSpot-Pro (a 2D GUI-grounding setting) and MMMU (STEM-heavy multimodal reasoning). This matches the intuition that depth-centric supervision is less aligned with those evaluation demands.

The right panel quantifies cross-task divergence per benchmark.
A larger top-to-bottom accuracy span indicates that performance is more sensitive to which synthetic task is used for training.
This reveals both task-dependent and benchmark-dependent transfer behavior.
Even among visual perception benchmarks, response profiles differ. RealWorldQA shows small divergence across tasks because it places stronger emphasis on reasoning. CV-Bench responds more strongly to structured tasks.

Overall, these analyses indicate that synthetic data can play a targeted improvement role when aligned with specific capabilities and can complement natural data in our current setting.
Expanding task diversity may unlock gains across a broader range of benchmarks.

\section{Related Work}
\label{sec:related_work}

\paragraph{Synthetic data for VLMs.}
Recent work increasingly uses synthetic multimodal data to scale vision--language learning and instruction tuning.
For caption generation, methods generate high-quality captions at scale: ShareGPT4V produces 1.2M captions \citep{chen2024sharegpt4v}, ALLaVA synthesizes 3.4M captioning and reasoning QA pairs \citep{chen2024allava}, SynthVLM refines caption quality \citep{liu2024synthvlm}, and VILA$^2$ enables self-recaptioning \citep{fang2024vila2}.
For instruction tuning, SVIT scales to 4.2M instructions \citep{zhao2023svit}, ProVision generates 10M+ vision-centric instructions \citep{zhang2025provision}, MMEvol evolves 447K instructions \citep{luo2025mmevol}, LOVA3 teaches VQA generation \citep{zhao2024lova3}, and Img-Diff creates contrastive pairs \citep{jiao2024imgdiff}.
Domain-specific methods include SpatialVLM for 3D reasoning \citep{chen2024spatialvlm}, Math-LLaVA for mathematical reasoning \citep{shi2024mathllava}, and chart QA pipelines \citep{yang2025ecd}.
Other work explores compositional generalization \citep{li2025sparcl}, failure-driven synthesis \citep{stan2025reasonfailures,wu2025compact}, and robustness analysis \citep{hu2025modelcollapse}.
\dataengine differs from prior work in being a fully automated pipeline that generates datasets from task keywords alone, while many related methods operate at a different scope and are therefore not directly comparable.

\paragraph{T2I-generated training data.}
T2I-generated training data has evolved from representation learning to supervised recognition.
For representation learning, StableRep applies diffusion images to contrastive learning \citep{tian2023stablerep}, Synth$^2$ uses T2I embeddings for VLM training \citep{sharifzadeh2024synth2}, and scaling laws show CLIP pretraining behavior \citep{fan2024scalinglaws}.
For classification, work examines synthetic-only versus mixed training \citep{he2023syntheticready}, filtered diffusion samples \citep{azizi2023syntheticimagenet}, low-data regimes \citep{zhou2023thinair}, robustness interventions \citep{shipard2023diversityneeded,yuan2022notjustprettypictures}, and generation-selection loops \citep{dunlap2023diversifyvision}.
For dense prediction, methods generate structured annotations \citep{wu2023datasetdm}, segmentation masks \citep{nguyen2023datasetdiffusion,wu2023diffumask}, and detection data with geometric control \citep{ni2023imaginarynet,chen2023geodiffusion}.
Beyond recognition, diffusion samples mitigate catastrophic forgetting \citep{wu2025gift}, while benchmarks support hybrid synthetic+natural recipes \citep{singh2024syntheticrobustness}.

\paragraph{VLM training and evaluation.}
Progress in visual instruction tuning relies on scalable supervision and evaluation.
Early pipelines, including LLaVA~\citep{liu2023llava}, InstructBLIP~\citep{dai2023instructblip}, MiniGPT-4~\citep{zhu2023minigpt4}, and Qwen-VL~\citep{bai2023qwenvl}, demonstrate that converting image--text corpora into instruction-following data aligns perception and generation in VLMs.
Visual perception benchmarks repurpose CV resources: CV-Bench~\citep{tong2024cambrian} uses Microsoft COCO~\citep{lin2014microsoft}, ADE20K~\citep{zhou2017sceneparsing}, and Omni3D~\citep{brazil2023omni3d} for 2D/3D perception, while RealWorldQA~\citep{xai2024realworldqa_dataset} targets real-scene spatial understanding.
Complementary benchmarks cover visual pitfalls such as MMVP~\citep{tong2024mmvp}, perception limitations such as BLINK~\citep{fu2024blink}, expert reasoning benchmarks such as MMMU~\citep{yue2023mmmu}, and integrated-competence benchmarks such as MM-Vet~\citep{yu2023mmvet} and MME~\citep{fu2023mme}.
Multi-domain benchmarks include MMBench~\citep{liu2023mmbench}, MMStar~\citep{chen2024mmstar}, MathVista~\citep{lu2024mathvista}, ScreenSpot-Pro~\citep{li2025screenspotpro}, and OCRBench~\citep{liu2023ocrbench}.
These resources help diagnose limitations and motivate targeted synthesis strategies.

\section{Conclusions and Discussion}
\label{sec:conclusion}

\paragraph{Limitations and future work.}
\datasetname focuses primarily on visual perception tasks, and our study investigates whether targeted synthetic supervision can improve capabilities such as spatial understanding, attribute recognition, and viewpoint reasoning.
An open question is whether a synthetic data pipeline such as \dataengine can naturally extend to more complex visual reasoning tasks that require longer inference chains and stronger compositional reasoning.
We leave this direction to future work and hope our findings motivate further study.

\paragraph{Conclusions.}
We present \dataengine, a synthetic data generation pipeline that uses an LLM to produce questions, answers, and T2I prompts, synthesizes images with T2I models, and applies multimodal alignment verification, all without reference images or human annotation.
Using \dataengine, we construct \datasetname, a VQA dataset containing 10k image--question--answer triples spanning 10 visual perception tasks.
Experiments across three VLMs show that training on \datasetname yields consistent improvements on visual perception benchmarks while preserving general multimodal capabilities.
The same targeted supervision also remains effective under RL, and a self-verified open-source pipeline continues to improve visual perception.

Our findings indicate that the perception bottleneck in VLMs is partly a data problem and that synthetic data can serve as a complementary source of supervision for strengthening core perception capabilities.
Progress in multimodal systems may depend not only on scaling model size or raw data volume, but also on constructing targeted supervision for perceptual distinctions that current models underlearn.
An automated synthetic data pipeline offers a practical and scalable building block for more systematic multimodal training.

\section*{Acknowledgments}
We thank Linrong Cai for his helpful review. We gratefully acknowledge the use of the Neuronic GPU computing cluster maintained by the Department of Computer Science at Princeton University. 

\bibliographystyle{plainnat}
\bibliography{visionfoundry}

\clearpage

\appendix

\raggedbottom
\renewcommand{\theHsection}{supp.\Alph{section}}
\renewcommand{\theHsubsection}{supp.\Alph{section}.\arabic{subsection}}

\thispagestyle{plain}
{\noindent\fontsize{24}{29}\selectfont\textbf{Appendix}\par}
\vspace{0.2em}


\vspace{0.25em}
\begin{itemize}[leftmargin=1.15em,itemsep=0.3em,topsep=0.15em,parsep=0pt,partopsep=0pt]
    \item \hyperref[app:impl_details]{\textcolor{appendixaccent}{\ref*{app:impl_details}}} presents optimization hyperparameters, evaluation protocol, and supplementary training configurations.
    \item \hyperref[app:exact_prompts]{\textcolor{appendixaccent}{\ref*{app:exact_prompts}}} gives the exact prompt sets used for sample generation and verification.
    \item \hyperref[app:case_studies]{\textcolor{appendixaccent}{\ref*{app:case_studies}}} collects representative verifier case studies across correct and incorrect outcomes.
    \item \hyperref[app:data_quality]{\textcolor{appendixaccent}{\ref*{app:data_quality}}} reports data-quality evidence through verification records and confusion statistics.
    \item \hyperref[app:taskwise_results]{\textcolor{appendixaccent}{\ref*{app:taskwise_results}}} lists task-wise settings and complete per-task benchmark results.
    \item \hyperref[app:verification_ablation]{\textcolor{appendixaccent}{\ref*{app:verification_ablation}}} studies the helpfulness of verification under a matched training budget.
    \item \hyperref[app:natural_vs_synthetic]{\textcolor{appendixaccent}{\ref*{app:natural_vs_synthetic}}} compares natural-image and synthetic-image supervision under controlled settings.
    \item \hyperref[app:opensource_t2i]{\textcolor{appendixaccent}{\ref*{app:opensource_t2i}}} evaluates an open-source T2I model.
    \item \hyperref[app:opensource_verifier]{\textcolor{appendixaccent}{\ref*{app:opensource_verifier}}} evaluates verifier choice and full-scale self-verification.
\end{itemize}

\begingroup
\hypersetup{linkcolor=black}
\etoctoccontentsline{part}{}
\etocstandardlines
\etocsettocstyle{}{}
\etocsetnexttocdepth{subsection}
\localtableofcontents
\endgroup

\clearpage

\section{Implementation Details}
\label{app:impl_details}
\smallskip
\subsection{Optimization and Hyperparameters}
\label{app:opt_details}
Table~\ref{tab:hyperparams} summarizes the model-specific hyperparameters used in the main experiments.
We report settings for Qwen2.5-VL-3B-Instruct~\citep{bai2025qwen25vl}, MiMo-VL-7B-SFT~\citep{coreteam2025mimovltechnicalreport}, and Llama-3.2-11B-Vision-Instruct~\citep{meta2024llama32vision,dubey2024llama3}.
In particular, $\eta_{\text{ViT}}$, $\eta_{\text{adapter}}$, and $\eta_{\text{LLM}}$ denote the learning rates for the ViT encoder, adapter modules, and LLM backbone, respectively.
All main-experiment runs use a global batch size of 128 and are trained for one epoch.

\begin{table}[H]
\centering
\small
\resizebox{\linewidth}{!}{%
\begin{tabular}{lcccccc}
\toprule
\textbf{Model} & \textbf{Epochs} & \textbf{Batch} & $\eta_{\text{ViT}}$ & $\eta_{\text{adapter}}$ & $\eta_{\text{LLM}}$ & \textbf{LLM} \\
\midrule
Qwen2.5-VL-3B-Instruct & 1 & 128 & $5\times10^{-7}$ & $5\times10^{-6}$ & $5\times10^{-6}$ & Unfrozen \\
MiMo-VL-7B-SFT & 1 & 128 & $5\times10^{-7}$ & $2.5\times10^{-6}$ & $2.5\times10^{-6}$ & Unfrozen \\
Llama-3.2-11B-Vision-Instruct & 1 & 128 & $5\times10^{-7}$ & $5\times10^{-6}$ & N/A & Frozen \\
\bottomrule
\end{tabular}%
}
\caption{\textbf{Main-experiment hyperparameters.} Training settings for Qwen2.5-VL-3B-Instruct, MiMo-VL-7B-SFT, and Llama-3.2-11B-Vision-Instruct in the primary comparison (all runs use 1 epoch and global batch size 128). $\eta_{\text{ViT}}$, $\eta_{\text{adapter}}$, and $\eta_{\text{LLM}}$ denote learning rates for the vision encoder, adapter modules, and LLM backbone, respectively; \textbf{LLM} indicates whether the LLM backbone is frozen or unfrozen.}
\label{tab:hyperparams}
\end{table}

For MiMo-VL-7B-SFT~\citep{coreteam2025mimovltechnicalreport}, we target the non-thinking setting during both training and evaluation.
During training, we prepend \texttt{<think></think>} to each answer as a placeholder to indicate disabled thinking.
During evaluation, we append \texttt{/no\_think} to the end of the input, following the official repository protocol, to measure visual perception gains more directly, rather than relying on indirect gains from extended reasoning\footnote{\url{https://github.com/XiaomiMiMo/MiMo-VL}}.
For Llama-3.2-11B-Vision-Instruct~\citep{meta2024llama32vision,dubey2024llama3}, we freeze the LLM backbone and only unfreeze ViT and adapters to mitigate catastrophic forgetting caused by full-parameter finetuning.

\subsection{Evaluation Protocol}
\label{app:eval_protocol}
All evaluations are conducted using the open-source VLMEvalKit framework~\citep{duan2024vlmevalkit} with default settings unless explicitly noted below.
For Qwen2.5-VL-3B-Instruct, we keep VLMEvalKit defaults and do not impose explicit minimum or maximum pixel overrides.

\paragraph{Benchmark suite.}
We evaluate models on 13 benchmarks in three categories.
Visual perception benchmarks include MMVP~\citep{tong2024mmvp}, CV-Bench~\citep{tong2024cambrian}, and RealWorldQA~\citep{xai2024realworldqa_dataset}.
General reasoning benchmarks include BLINK~\citep{fu2024blink}, MMMU-Val~\citep{yue2023mmmu}, MMBench (EN v1.1)~\citep{liu2023mmbench}, MMStar~\citep{chen2024mmstar}, and MathVista-Mini~\citep{lu2024mathvista}.
Application-specific benchmarks include OCRBench~\citep{liu2023ocrbench}, ScreenSpot-Pro~\citep{li2025screenspotpro}, MMSI-Bench (circular)~\citep{yang2025mmsibench}, 3DSRBench~\citep{ma2025_3dsrbench}, and LEGO (circular)~\citep{tang2025legopuzzles}.

\paragraph{Benchmark-specific settings.}
For MMVP, we report both the official pair accuracy and the commonly used single-image accuracy for reference.
For MMMU, we evaluate on the \texttt{val} split only and exclude the \texttt{dev} split.
For ScreenSpot-Pro, we report the average of overall scores across all subtasks.
For LEGO and MMSI-Bench, we use the default circular evaluation mode in VLMEvalKit.
For 3DSRBench, we use VLMEvalKit's default FlipEval result in circular-evaluation mode (the most challenging mode).
All reported numbers are averaged over four independent runs. All reported deltas are computed before rounding; absolute scores are rounded to one decimal place.

\subsection{Supplementary Protocols}
\label{app:supp_protocols}
\paragraph{Data-size sensitivity.}
We reuse exactly the same model hyperparameters as the main experiments.
Subsets are built by random sampling from the full dataset at sizes 0.5k, 1k, 2k, and 5k.
For each subset size, results are averaged over four independent runs; each run is trained for one epoch, and evaluation settings are identical to the main experiments.

\paragraph{Synthetic vs.\ natural data.}
We keep model hyperparameters unchanged from the main experiments.
The synthetic set is a random subset sampled from \datasetname, and the natural set is a random subset sampled from LLaVA-Instruct-80K~\citep{liu2023llava}, with matched subset sizes.
Results are averaged over four independent runs, each trained for one epoch with the same evaluation protocol as the main experiments.

\paragraph{Training epochs and task-wise settings.}
For the training-epoch ablation, we use a randomly selected 1k subset and vary only the number of epochs, while keeping training and evaluation settings identical to the main experiments.
For task-wise experiments, we train one model per task-specific dataset for one epoch under the same training and evaluation settings as the main experiments.

\subsection{RL Settings}
\label{app:rl_details}
For all RL experiments, we use Group Relative Policy Optimization (GRPO)~\citep{shao2024deepseekmath} with four rollouts per prompt and a KL divergence penalty coefficient of 0.01. We use the same 10-task 10k synthetic dataset (\datasetname) as in the main experiments and train for one epoch. Model-specific learning rates, batch sizes, and trainable modules are summarized in Table~\ref{tab:rl_hyperparams}. The reward model (Qwen2.5-3B) serves as a language judge to evaluate rollout responses against the ground-truth answers, providing a binary correctness reward.

\begin{table}[H]
\centering
\small
\resizebox{\linewidth}{!}{%
\begin{tabular}{lccccccl}
\toprule
\textbf{Model} & \textbf{Trainable modules} & \textbf{Reward model} & \textbf{Epoch} & \textbf{LR} & \textbf{Batch} & \textbf{Rollouts} & \textbf{Details} \\
\midrule
Qwen2.5-VL-3B-Instruct & Full model & Qwen2.5-3B & 1 & $4\times10^{-6}$ & 128 & 4 & GRPO, step 70 \\
Llama-3.2-11B-Vision-Instruct & Vision + adapters & Qwen2.5-3B & 1 & $1\times10^{-6}$ & 32 & 4 & GRPO, step 281 \\
MiMo-VL-7B-SFT & Full model & Qwen2.5-3B & 1 & $1\times10^{-6}$ & 64 & 4 & GRPO, step 140 \\
\bottomrule
\end{tabular}%
}
\caption{\textbf{RL hyperparameters.} Experimental configurations for GRPO post-training on \datasetname. All runs use a KL coefficient of 0.01 and four rollouts per prompt.}
\label{tab:rl_hyperparams}
\end{table}

Across vision benchmarks, RL post-training consistently maintains or improves model performance across backbones, confirming that \dataengine supervision transfers effectively beyond SFT.

\section{Exact Pipeline Prompts}
\label{app:exact_prompts}
We report the exact prompt templates used in \dataengine\ for generation, verification, and feedback.
Fields in angle brackets (e.g., \texttt{<TASK\_DESCRIPTION>}) denote runtime substitutions from
\texttt{TaskConfig} or states.

\subsection{Prompt Set A}
\label{app:prompts_generation}

\begin{promptstylebox}{A1. Custom attribute expansion (e.g., spatial relations, difficulty labels).}
\noindent\textbf{System prompt:}
\begin{verbatim}
Output ONLY valid JSON array of strings.
\end{verbatim}

\noindent\textbf{User prompt:}
\begin{verbatim}
For the task '<TASK_DESCRIPTION>', generate <COUNT> diverse options for '<ATTR_NAME>'.
These should be specific, varied, and visually verifiable if possible.
Return ONLY a valid JSON array of strings.
\end{verbatim}
\end{promptstylebox}

This prompt expands task attributes into a controlled candidate pool that is later used to instantiate concrete samples.
The JSON-only requirement keeps outputs machine-readable, while the diversity and visual-verifiability constraints encourage attributes that are both varied and practically checkable during downstream generation and verification.

\begin{promptstylebox}{A2. Single-image QA + T2I case construction.}
\noindent\textbf{System prompt template:}
\begin{verbatim}
You are an expert VQA data creator for the task: "<TASK_ID>"
Task description: <TASK_DESCRIPTION>

<IF NUM_OBJECTS > 1>
This task requires <NUM_OBJECTS> objects in the image with clear relationships.

<IF MODE == "multi">
MULTI-IMAGE MODE: Generate <NUM_IMAGES> interconnected images with a coherent story.

<IF "spatial" IN TASK_ID OR CONSTRAINTS>
SPATIAL FOCUS: The question MUST involve spatial relationships between objects.

<IF "color" IN TASK_ID OR CONSTRAINTS>
COLOR FOCUS: Object colors must be distinct, accurate, and central to the question.

CRITICAL RULES:
1. The answer-determining fact MUST be 100% visually verifiable from the final image.
2. Text prompt must explicitly describe content matching the correct answer.
3. Never rely on invisible properties.
4. Generate deterministic, unambiguous questions.

Constraints: <COMMA_JOINED_CONSTRAINTS_OR_NONE>
Return EXACTLY ONE JSON object with keys:
- "prompt": extremely detailed text-to-image prompt (English)
- "question": clear VQA question
- "answer": clear deterministic answer
- "metadata": {"difficulty": "easy", "category": "<TASK_ID>", "num_objects": <NUM_OBJECTS>}
\end{verbatim}

\end{promptstylebox}

\clearpage

This prompt is the main sample-construction template: it jointly produces the image prompt, question, and answer in one structured record.
Conditional blocks (e.g., spatial/color focus and object-count constraints) adapt instructions to task requirements, and the required rules explicitly enforce visual grounding and determinism so that each generated case remains auditable and suitable for training.

\subsection{Prompt Set B}
\label{app:prompts_verification}

\begin{promptstylebox}{B1. Question--answer to declarative verification statement.}
\noindent\textbf{System prompt:}
\begin{verbatim}
Rewrite question+answer into ONE concise sentence. Output only the sentence.
\end{verbatim}

\noindent\textbf{User prompt:}
\begin{verbatim}
Q: <QUESTION>
A: <ANSWER>
\end{verbatim}

\end{promptstylebox}

This prompt standardizes each QA pair into a single declarative statement that can be checked directly against visual evidence.
By removing interrogative wording and enforcing one concise sentence, it reduces ambiguity for downstream verification and keeps the judged content aligned with the original answer target.

\begin{promptstylebox}{B2. Image consistency verification (binary judge).}
\noindent\textbf{System prompt:}
\begin{verbatim}
You are an image verification assistant. Decide if an image matches the statement overall,
ignoring minor details. ANSWER WITH EXACTLY ONE LINE: 'Answer: YES' or 'Answer: NO'
\end{verbatim}

\noindent\textbf{User payload text field:}
\begin{verbatim}
Statement: '<STATEMENT>'

Does this image match?
\end{verbatim}

\end{promptstylebox}

This prompt performs the core filtering decision: whether the generated image supports the target statement at an overall semantic level.
The instruction to ignore minor details makes the check robust to inconsequential rendering noise, while the strict one-line \texttt{YES/NO} format enables deterministic parsing in the data pipeline.
The verifier receives the image in the same request (URL or base64 data URI), and only \texttt{Answer: YES} counts.

\begin{promptstylebox}{B3. Optional edit-based regeneration after failed verification.}
\noindent\textbf{Initial image-generation instruction:}
\begin{verbatim}
generate an image: <PROMPT>
\end{verbatim}

\noindent\textbf{Edit instruction (used when \texttt{use\_edit=True}):}
\begin{verbatim}
Make the image exactly match: '<STATEMENT>'. Only modify what's needed.
\end{verbatim}

\end{promptstylebox}
This prompt defines the fallback correction stage for cases that fail verification.
The first instruction generates an image from the original prompt, and the edit instruction then applies minimal, targeted changes to satisfy the verification statement without unnecessary drift.
Operationally, each sample uses outer-loop retries up to the configured maximum; each verification call has retry handling; and retention requires approval.

\clearpage

\section{Verifier Cases}
\label{app:case_studies}
We provide qualitative case studies to interpret how synthetic samples are handled by the verifier.
Following a manual audit, we classify each sample along two dimensions: (i) whether the generation is correct (correct/wrong), and (ii) whether the verifier's judgment is correct.
We then report the filter's final decision (accept/reject): \textit{true positive} (correct and accepted), \textit{false negative}
(correct but rejected), \textit{false positive} (wrong but accepted), and \textit{true negative} (wrong and rejected).

\subsection{True Positive Cases}
\label{app:case_pass}
We present representative examples where generation is correct and the verifier accepts the sample.

\begin{tcolorbox}[
  enhanced,
  colback=green!3,
  colframe=green!35!black,
  boxrule=0.9pt,
  sharp corners,
  title={Case 1: surface glossiness judgment},
  fonttitle=\sffamily\fontseries{bx}\selectfont,
  fontupper=\small\sffamily,
  left=10pt, right=10pt, top=8pt, bottom=8pt,
  toptitle=5pt, bottomtitle=5pt, lefttitle=10pt,
  boxsep=0pt
]
\begin{minipage}[t]{0.28\linewidth}
\vspace{0pt}
\vspace{0pt}
\centering
\includegraphics[width=\linewidth]{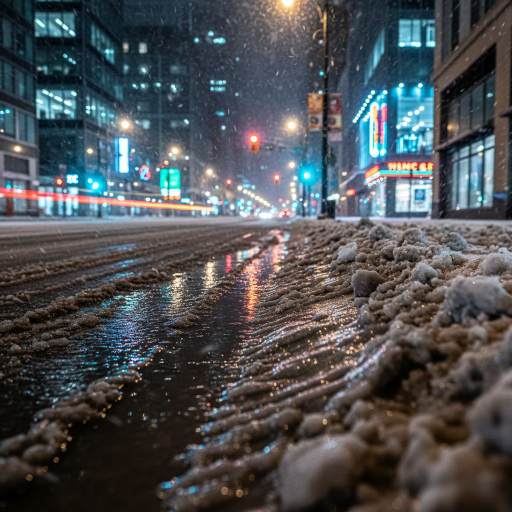}
\end{minipage}
\hfill
\begin{minipage}[t]{0.68\linewidth}
\vspace{0pt}
\raggedright
\vspace{0pt}
\raggedright
\small
\textbf{Question:} Does the snowbank in the foreground appear glossy (wet and reflective)?\\
\vspace{15pt}
\textbf{Answer:} Yes, it appears glossy.
\end{minipage}
\end{tcolorbox}

\begin{tcolorbox}[
  enhanced,
  colback=green!3,
  colframe=green!35!black,
  boxrule=0.9pt,
  sharp corners,
  title={Case 2: relative depth comparison},
  fonttitle=\sffamily\fontseries{bx}\selectfont,
  fontupper=\small\sffamily,
  left=10pt, right=10pt, top=8pt, bottom=8pt,
  toptitle=5pt, bottomtitle=5pt, lefttitle=10pt,
  boxsep=0pt
]
\begin{minipage}[t]{0.28\linewidth}
\vspace{0pt}
\centering
\includegraphics[width=\linewidth]{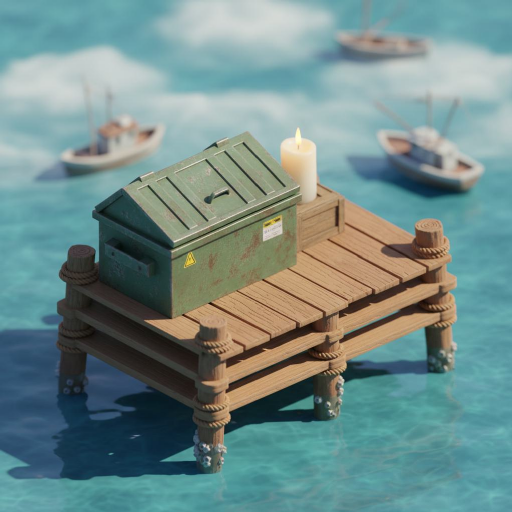}
\end{minipage}
\hfill
\begin{minipage}[t]{0.68\linewidth}
\vspace{0pt}
\raggedright
\small
\textbf{Question:} In the scene, which object is closer to the camera: the dumpster or the candle?\\
\vspace{15pt}
\textbf{Answer:} The dumpster is closer to the camera.
\end{minipage}
\end{tcolorbox}

\begin{tcolorbox}[
  enhanced,
  colback=green!3,
  colframe=green!35!black,
  boxrule=0.9pt,
  sharp corners,
  title={Case 3: high-viewpoint recognition},
  fonttitle=\sffamily\fontseries{bx}\selectfont,
  fontupper=\small\sffamily,
  left=10pt, right=10pt, top=8pt, bottom=8pt,
  toptitle=5pt, bottomtitle=5pt, lefttitle=10pt,
  boxsep=0pt
]
\begin{minipage}[t]{0.28\linewidth}
\vspace{0pt}
\centering
\includegraphics[width=\linewidth]{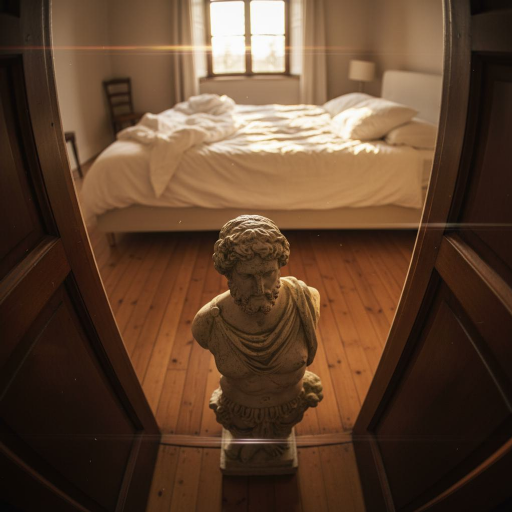}
\end{minipage}
\hfill
\begin{minipage}[t]{0.68\linewidth}
\vspace{0pt}
\raggedright
\small
\textbf{Question:} Is the photo taken from above the statue looking downward, as if from a balcony-like high viewpoint?\\
\vspace{15pt}
\textbf{Answer:} Yes, it is taken from above looking downward.
\end{minipage}
\end{tcolorbox}

\clearpage

\subsection{False Negative Cases}
\label{app:case_correct_fail}
We present a false-negative case in which generation is correct but the verifier rejects the sample.

\begin{tcolorbox}[
  enhanced,
  colback=yellow!6,
  colframe=orange!50!black,
  boxrule=0.9pt,
  sharp corners,
  title={Case 4},
  fonttitle=\sffamily\fontseries{bx}\selectfont,
  fontupper=\small\sffamily,
  left=10pt, right=10pt, top=8pt, bottom=8pt,
  toptitle=5pt, bottomtitle=5pt, lefttitle=10pt,
  boxsep=0pt
]
\begin{minipage}[t]{0.28\linewidth}
\vspace{0pt}
\centering
\includegraphics[width=\linewidth]{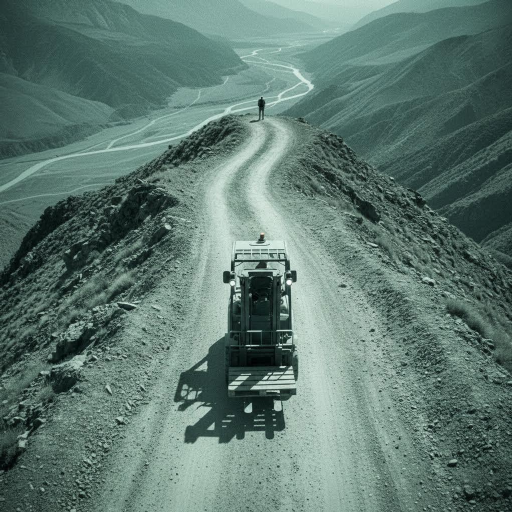}
\end{minipage}
\hfill
\begin{minipage}[t]{0.68\linewidth}
\vspace{0pt}
\raggedright
\small
\textbf{Statement:} The forklift is facing toward the camera.\\
\vspace{15pt}
\textbf{Reason:} The image matches the statement (the forklift indeed faces toward the camera), but the VLM verifier incorrectly judges it as wrong, leading to a false reject.
\end{minipage}
\end{tcolorbox}

\subsection{False Positive Cases}
\label{app:case_wrong_fail}
We present a false-positive case in which generation is wrong but the verifier accepts the sample.

\begin{tcolorbox}[
  enhanced,
  colback=orange!6,
  colframe=orange!60!black,
  boxrule=0.9pt,
  sharp corners,
  title={Case 5},
  fonttitle=\sffamily\fontseries{bx}\selectfont,
  fontupper=\small\sffamily,
  left=10pt, right=10pt, top=8pt, bottom=8pt,
  toptitle=5pt, bottomtitle=5pt, lefttitle=10pt,
  boxsep=0pt
]
\begin{minipage}[t]{0.28\linewidth}
\vspace{0pt}
\centering
\includegraphics[width=\linewidth]{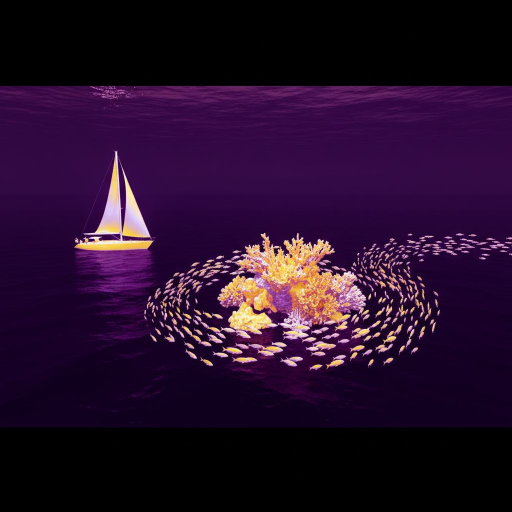}
\end{minipage}
\hfill
\begin{minipage}[t]{0.68\linewidth}
\vspace{0pt}
\raggedright
\small
\textbf{Statement:} The sailboat's bow is pointing left.\\
\vspace{15pt}
\textbf{Reason:} The generated image is incorrect: the sailboat's bow points right instead of left. However, the VLM verifier mistakenly accepts this sample, resulting in a false accept.
\end{minipage}
\end{tcolorbox}

\subsection{True Negative Cases}
\label{app:case_wrong_pass}
We present a true-negative case in which generation is wrong and the verifier rejects the sample.

\begin{tcolorbox}[
  enhanced,
  colback=red!3,
  colframe=red!45!black,
  boxrule=0.9pt,
  sharp corners,
  title={Case 6},
  fonttitle=\sffamily\fontseries{bx}\selectfont,
  fontupper=\small\sffamily,
  left=10pt, right=10pt, top=8pt, bottom=8pt,
  toptitle=5pt, bottomtitle=5pt, lefttitle=10pt,
  boxsep=0pt
]
\begin{minipage}[t]{0.28\linewidth}
\vspace{0pt}
\centering
\includegraphics[width=\linewidth]{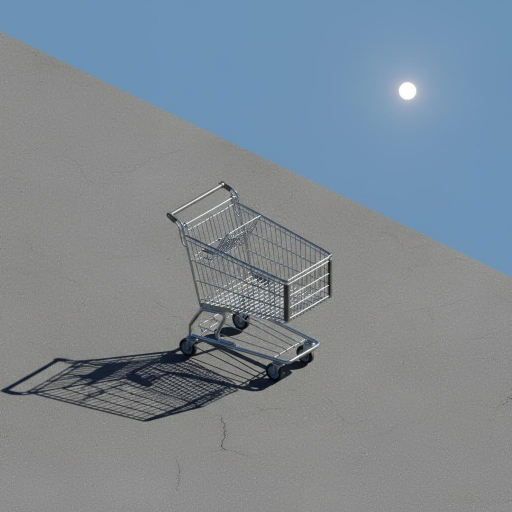}
\end{minipage}
\hfill
\begin{minipage}[t]{0.68\linewidth}
\vspace{0pt}
\raggedright
\small
\textbf{Statement:} The shopping cart is facing toward the lower-left corner of the scene.\\
\vspace{15pt}
\textbf{Reason:} The generation is wrong: it should face toward the lower-left corner but is rendered toward the lower-right. The VLM verifier correctly identifies this mismatch and rejects the sample.
\end{minipage}
\end{tcolorbox}

\clearpage

\section{Data Quality Analysis}
\label{app:data_quality}
To quantify the reliability of generated supervision, we report the acceptance dynamics of our filtering pipeline and manual agreement statistics.
This section analyzes one concrete random-sampling run in which we track the full generation trajectory to obtain the final valid subset.
These diagnostics show whether high-quality supervision emerges systematically rather than by chance.

\subsection{Verification-Accuracy Records}
\label{app:verification_records}

\begin{wrapfigure}{l}{0.37\linewidth}
\centering
\includegraphics[width=0.97\linewidth]{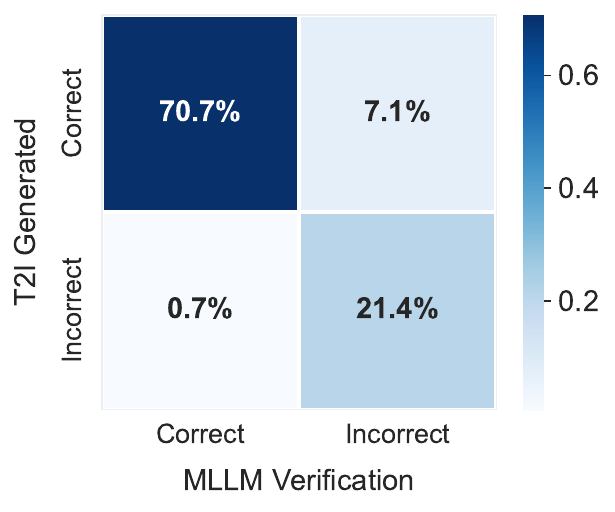}
\caption{\textbf{Confusion matrix for the random-sampling audit.} Rows are manual correctness labels, and columns are verifier outputs. Values summarize the audited trajectory.}
\label{fig:verification_confusion_matrix}
\end{wrapfigure}
We summarize the audited trajectory with a confusion matrix.
We set a target of retaining 100 samples and therefore audited one full production trajectory with 140 generation attempts, in which the verifier filtered out 40 candidates.
Here, a valid case is defined as \textit{generated correct \& filter pass} under manual audit.
To avoid survivorship bias, we audit the full trajectory rather than only the retained outputs.
The resulting outcome shares are: generated correct \& filter pass = 70.7\%,
generated correct \& filter fail = 7.1\%, generated wrong \& filter pass = 0.7\%, and
generated wrong \& filter fail = 21.4\%.
In absolute counts, this corresponds to 140 audited candidates with 100 retained after verification: 99 valid samples (generated correct \& filter pass) and 1 false accept (generated wrong \& filter pass); the remaining 40 are rejected (10 generated correct \& filter fail and 30 generated wrong \& filter fail).
As visualized in Figure~\ref{fig:verification_confusion_matrix}, most mass lies on the diagonal,
with \textit{correct \& pass} and \textit{wrong \& fail} as the dominant outcomes.
This indicates that the verifier is directionally aligned with manual judgments on this audited run,
effectively capturing the dominant correctness patterns in the generation stream.
Using generation correctness as ground truth and filter pass/fail as predictions,
the filter output rates are 71.4\% pass and 28.6\% fail,
and the filter-pass set is highly reliable (99.0\% correct).

\begin{table}[H]
\centering
\small
\begin{tabularx}{\linewidth}{l*{4}{Y}}
\toprule
\textbf{Split} & \textbf{Accuracy} & \textbf{Precision} & \textbf{Recall} & \textbf{F1} \\
\midrule
Overall audited set & 92.1 & 99.0 & 90.8 & 94.7 \\
\bottomrule
\end{tabularx}
\caption{\textbf{Verifier performance on the audited sample set.} Accuracy, precision, recall, and F1 are computed by comparing verifier pass/fail outputs against manual labels over all audited candidates.}
\label{tab:verification_records}
\end{table}
\vspace{-10pt}
Table~\ref{tab:verification_records} shows high precision (99.0\%) and strong F1 (94.7\%),
suggesting that accepted samples are typically correct under manual audit.
Recall (90.8\%) is lower than precision, indicating that some correct samples are filtered out.
Overall, the verifier behaves as a high-precision filter in this trajectory.
The audit reflects the automated filtering distribution without manual post-processing, characterizing the raw data quality yielded by the verifier.

\begin{table}[H]
\centering
\small
\resizebox{0.97\linewidth}{!}{%
\begin{tabular}{lcc}
\toprule
\textbf{Metric} & \textbf{Definition} & \textbf{Value} \\
\midrule
Final valid proportion & generated correct \& filter pass over all audited candidates & 70.7\% \\
Acceptance rate & retained / generated candidates & 71.4\% \\
Rejection rate & rejected / generated candidates & 28.6\% \\
False accept rate & generated wrong but filter pass / generated wrong candidates & 3.2\% \\
False reject rate & generated correct but filter fail / generated correct candidates & 9.2\% \\
Human--judge agreement & Cohen's $\kappa$ on audited subset & 0.794 \\
\bottomrule
\end{tabular}
}
\caption{\textbf{Pipeline quality metrics on the audited trajectory.} The table reports final valid proportion, acceptance/rejection rates, false accept/reject rates, and human--judge agreement (Cohen's $\kappa$).}
\label{tab:data_quality}
\end{table}

Table~\ref{tab:data_quality} provides an operational view: the pipeline retains 71.4\% of candidates,
which in this audit is exactly 100/140 retained (99 valid and 1 false accept) and 40/140 rejected.
This is equivalent to a final valid proportion of 70.7\% and low false accepts (3.2\%).
False rejects remain non-zero (9.2\%), showing a measurable quality--coverage trade-off.
The observed Cohen's $\kappa$ of 0.794 is consistent with substantial agreement,
but this estimate is based on one audited trajectory and should be interpreted as indicative rather than universal.
\vspace{-5pt}
\section{Task-wise Results}
\label{app:taskwise_results}
This section reports full task-wise experimental settings and complete per-task benchmark results.
\vspace{-5pt}
\subsection{Task-wise Experimental Settings}
\label{app:taskwise_settings}
Each task-specific dataset is trained for one epoch.
Training and evaluation settings are identical to those used in the main experiments (Appendix~\ref{app:opt_details} and \ref{app:eval_protocol}).
This ensures that performance differences are attributable to task specialization rather than variations in evaluation protocols.
\subsection{Complete Per-task Results}
\label{app:taskwise_complete_results}
Table~\ref{tab:taskwise_complete_part1} and Table~\ref{tab:taskwise_complete_part2} report the full benchmark results of all ten task-specific models, all based on Qwen2.5-VL-3B-Instruct.
To keep the tables compact, we use task abbreviations:
\textbf{OaD} (\emph{Orientation and Direction}), \textbf{VaP} (\emph{Viewpoint and Perspective}), \textbf{PaRC} (\emph{Positional and Relational Context}), \textbf{SR} (\emph{Spatial Relationship}), \textbf{SaC} (\emph{State and Condition}), \textbf{SPC} (\emph{Structural and Physical Characteristics}), \textbf{CaA} (\emph{Color and Appearance}), \textbf{DO} (\emph{Depth Order}), \textbf{RD} (\emph{Relative Distance}), and \textbf{RWSU} (\emph{Real-World Spatial Understanding}).
Benchmark abbreviations are:
\textbf{MVP-S} (MMVP single-image), \textbf{MVP-P} (MMVP pair), \textbf{CV2D} (CV-Bench-2D), \textbf{CV3D} (CV-Bench-3D), \textbf{RWQA} (RealWorldQA), \textbf{BLK} (BLINK), \textbf{MMS} (MMStar), \textbf{OCR} (OCRBench), \textbf{MMMU} (MMMU-Val), \textbf{MMB} (MMBench), \textbf{MVis} (MathVista), \textbf{SSP} (ScreenSpot-Pro), \textbf{MMSI} (MMSI-Bench), and \textbf{3DSR} (3DSRBench).
Corresponding benchmark references are MMVP~\citep{tong2024mmvp}, CV-Bench~\citep{tong2024cambrian}, RealWorldQA~\citep{xai2024realworldqa_dataset}, BLINK~\citep{fu2024blink}, MMStar~\citep{chen2024mmstar}, OCRBench~\citep{liu2023ocrbench}, MMMU~\citep{yue2023mmmu}, MMBench~\citep{liu2023mmbench}, MathVista~\citep{lu2024mathvista}, ScreenSpot-Pro~\citep{li2025screenspotpro}, MMSI-Bench~\citep{yang2025mmsibench}, 3DSRBench~\citep{ma2025_3dsrbench}, and LEGO~\citep{tang2025legopuzzles}.

\begin{table}[H]
\centering
\scriptsize
\begin{tabularx}{\linewidth}{*{9}{Y}}
\toprule
\textbf{Task} & \textbf{MVP-S} & \textbf{MVP-P} & \textbf{CV2D} & \textbf{CV3D} & \textbf{RWQA} & \textbf{BLK} & \textbf{MMS} & \textbf{OCR} \\
\midrule
OaD & 64.0 & 36.0 & 68.4 & 73.9 & 66.4 & 49.1 & 55.8 & 82.4 \\
VaP & 65.0 & 36.0 & 68.4 & 73.4 & 65.6 & 48.9 & 55.8 & 81.8 \\
PaRC & 66.0 & 38.0 & 67.9 & 73.5 & 66.1 & 49.5 & 55.4 & 82.2 \\
SR & 64.3 & 36.0 & 67.9 & 72.7 & 66.1 & 49.0 & 55.3 & 82.3 \\
SaC & 64.7 & 36.7 & 68.3 & 71.5 & 65.6 & 48.9 & 55.4 & 82.0 \\
SPC & 65.0 & 36.0 & 67.9 & 73.7 & 66.0 & 48.5 & 55.7 & 82.4 \\
CaA & 64.7 & 36.7 & 68.6 & 73.7 & 65.9 & 48.5 & 56.1 & 82.2 \\
DO & 64.3 & 35.3 & 67.7 & 72.4 & 65.4 & 48.8 & 55.5 & 82.5 \\
RD & 63.3 & 34.7 & 67.5 & 74.0 & 66.3 & 48.9 & 55.9 & 82.3 \\
RWSU & 64.3 & 36.0 & 67.3 & 73.0 & 65.6 & 48.9 & 55.5 & 82.2 \\
\bottomrule
\end{tabularx}
\caption{\textbf{Task-wise benchmark performance across 10 task-specialized models (Part I).} Scores (\%) on MVP-S, MVP-P, CV2D, CV3D, RWQA, BLK, MMS, and OCR; higher is better.}
\label{tab:taskwise_complete_part1}
\end{table}
\begin{table}[H]
\centering
\scriptsize
\begin{tabularx}{\linewidth}{*{8}{Y}}
\toprule
\textbf{Task} & \textbf{LEGO} & \textbf{MMMU} & \textbf{MMB} & \textbf{MVis} & \textbf{SSP} & \textbf{MMSI} & \textbf{3DSR} \\
\midrule
OaD & 15.7 & 50.0 & 76.2 & 64.2 & 20.7 & 8.8 & 35.2 \\
VaP & 16.3 & 49.0 & 75.9 & 65.2 & 20.5 & 8.7 & 35.2 \\
PaRC & 16.4 & 48.8 & 76.8 & 64.6 & 20.6 & 9.4 & 35.3 \\
SR & 16.0 & 48.1 & 76.2 & 63.3 & 20.2 & 9.4 & 34.9 \\
SaC & 16.1 & 48.8 & 76.1 & 64.4 & 20.7 & 9.7 & 35.2 \\
SPC & 15.8 & 48.4 & 76.3 & 63.9 & 20.4 & 9.5 & 35.3 \\
CaA & 15.8 & 47.9 & 76.0 & 64.9 & 20.6 & 8.9 & 35.4 \\
DO & 15.8 & 46.6 & 76.2 & 63.6 & 18.7 & 9.1 & 35.3 \\
RD & 15.9 & 48.9 & 76.3 & 65.2 & 19.9 & 9.6 & 35.6 \\
RWSU & 16.0 & 48.2 & 76.2 & 64.2 & 20.6 & 9.1 & 35.2 \\
\bottomrule
\end{tabularx}
\caption{\textbf{Task-wise benchmark performance across 10 task-specialized models (Part II).} Scores (\%) on LEGO, MMMU, MMB, MVis, SSP, MMSI, and 3DSR; higher is better.}
\label{tab:taskwise_complete_part2}
\end{table}

\section{Ablation on Verification Helpfulness}
\label{app:verification_ablation}

To isolate the contribution of the verification stage in \dataengine, we conduct a controlled ablation on a single task with a fixed training size of 1k samples.

\paragraph{Goal and protocol.}
This ablation tests whether verification helps under a matched 1k-sample training budget.
We use Qwen2.5-VL-3B-Instruct as the backbone and evaluate three settings: (i) the baseline, (ii) finetuning with verification, and (iii) finetuning without verification.
For both finetuned models, we keep the training recipe identical: unfreeze adapters, ViT, and LLM; use learning rates of $5\times10^{-6}$ for adapter/LLM and $5\times10^{-7}$ for ViT; train for one epoch with batch size 128.
The training task is \emph{Orientation and Direction}. Both finetuned models use 1k training samples: the verified branch keeps 1k verifier-accepted samples, while the non-verified branch uses an unfiltered 1k-sample set from the same task recipe.

\begin{figure}[H]
    \centering
    \includegraphics[width=0.98\linewidth]{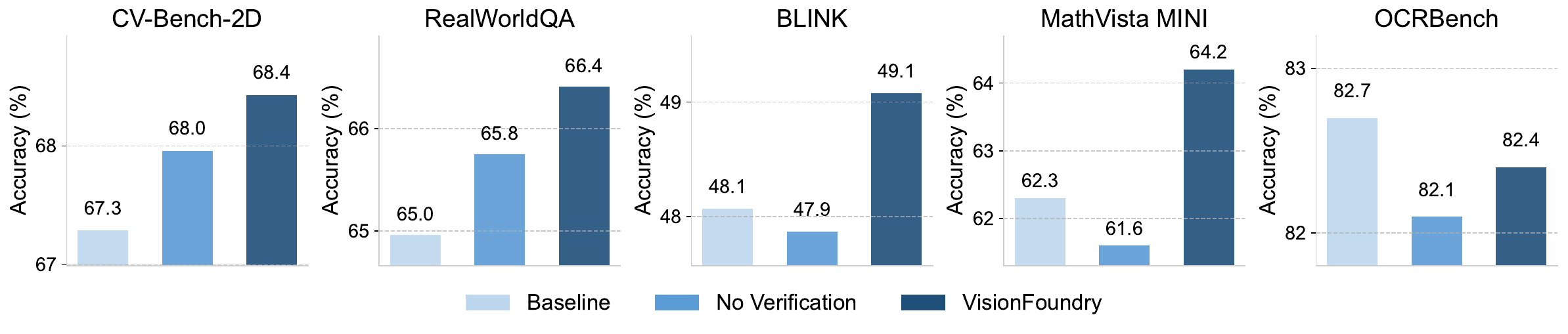}
    \caption{\textbf{Verification is helpful under a matched 1k-sample training budget.} The figure compares baseline, finetuning without verification, and finetuning with verification (\dataengine) on representative benchmarks. The verified branch performs better than the non-verified variant, with the clearest gains on visual perception evaluations.}
    \label{fig:verification_ablation_1k}
\end{figure}

\paragraph{Results and discussion.}
Figure~\ref{fig:verification_ablation_1k} shows that verification is helpful under a matched 1k-sample training budget.
Compared with \emph{without verification}, \emph{with verification} improves CV-Bench-2D by +0.5 points, RealWorldQA by +0.7 points, BLINK by +1.2 points, MathVista-Mini by +2.6 points, and OCRBench by +0.3 points.
Relative to the baseline, the verified run improves on four of five benchmarks (CV-Bench-2D, RealWorldQA, BLINK, MathVista-Mini), while the non-verified run falls below baseline on BLINK, MathVista-Mini, and OCRBench.
Overall, the gains are most visible on the vision-focused benchmarks listed above, while differences on other benchmarks are comparatively small. This confirms that verification is beneficial in this matched-1k setting.

\section{Synthetic Process Ablations}
\label{app:natural_vs_synthetic}

This section provides a practical comparison between a partially synthetic setting, real images + synthetic QA, and the full synthetic process used by \dataengine, synthetic images + synthetic QA. We defer stricter image-source isolation to the strict-control analysis.

\paragraph{Goal and protocol.}
The goal is to provide practical evidence on whether improvements mainly come from synthetic QA construction alone, or whether the full synthetic process itself provides additional training value.
In \dataengine, the T2I prompt, question, and answer are generated in the same dialogue turn, so synthetic images are naturally aligned with text supervision and do not require manual captioning.
For completeness, we construct a natural-image counterpart: we sample 1k images from LLaVA-Instruct-80K, generate captions with Gemini-3-Pro, then generate QA pairs from these captions and process them with the same pipeline to obtain a 1k training set.
We compare this dataset against the matched 1k synthetic \emph{Orientation and Direction} set under identical finetuning settings on Qwen2.5-VL-3B-Instruct (same unfreezing strategy, learning rates, epoch, and batch size), and report both against the no-finetuning baseline.

\begin{figure}[H]
    \centering
    \includegraphics[width=0.98\linewidth]{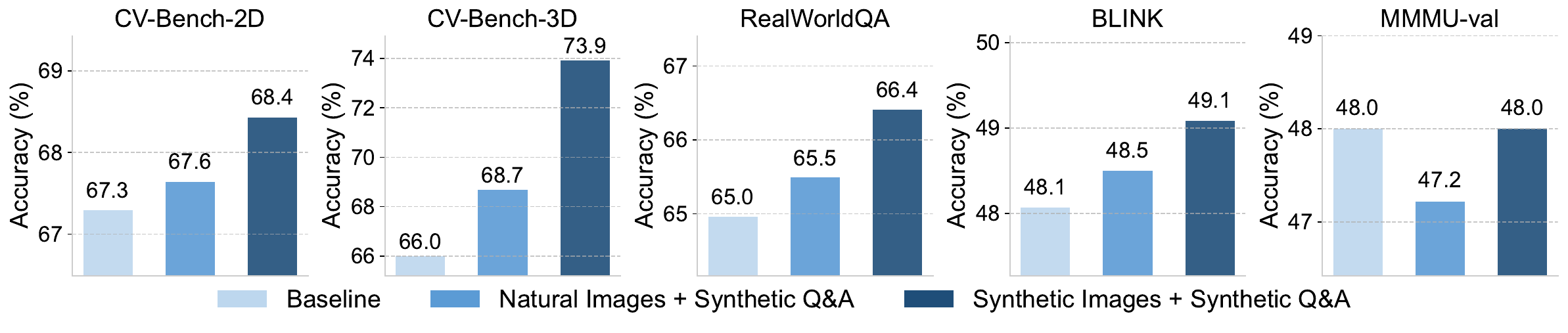}
    \caption{\textbf{Natural images with synthetic QA vs.\ synthetic images with synthetic QA under matched 1k-data training.} With the same model and setup, the synthetic-image setting shows stronger overall performance, with particularly clear gains on visual perception benchmarks while remaining comparable on the rest.}
    \label{fig:natural_vs_synthetic_1k}
\end{figure}

\paragraph{Results and discussion.}
Figure~\ref{fig:natural_vs_synthetic_1k} shows that the synthetic-image setting outperforms the natural-image setting on all five reported benchmarks.
Compared with real images + synthetic QA, synthetic images + synthetic QA improves CV-Bench-2D by +0.8 points, CV-Bench-3D by +5.3 points, RealWorldQA by +0.9 points, BLINK by +0.6 points, and MMMU-Val by +0.8 points.
Both settings are broadly competitive on general-purpose evaluation, while the synthetic-image setting provides clearly larger gains on visual perception metrics (especially CV-Bench-3D).
These results suggest that the full synthetic process provides added value beyond QA generation in \dataengine; we test this more rigorously in a strict-control analysis.

\paragraph{Strict-control ablation.}
To further isolate the image-source effect, we conduct a more strictly controlled variant where the synthetic QA component is explicitly matched across branches.
The natural-image branch follows
\emph{Natural Image $\rightarrow$ Gemini-3-Pro Caption $\rightarrow$ LLM Synthetic QA}.
For the synthetic-image branch, we reuse the same captions and QA, and only replace the image source via
\emph{Caption $\rightarrow$ T2I Model $\rightarrow$ Synthetic Image}.
Hence, the key controlled variable is the image source (natural vs.\ caption-conditioned synthetic reconstruction), while QA supervision is kept consistent.
The finetuning setup follows the controlled protocol established in Appendix~\ref{app:verification_ablation} for Qwen2.5-VL-3B-Instruct: unfreeze adapters, ViT, and LLM; learning rates of $5\times10^{-6}$ (adapter/LLM) and $5\times10^{-7}$ (ViT); one epoch; batch size 128.

\begin{figure}[H]
    \centering
    \includegraphics[width=0.98\linewidth]{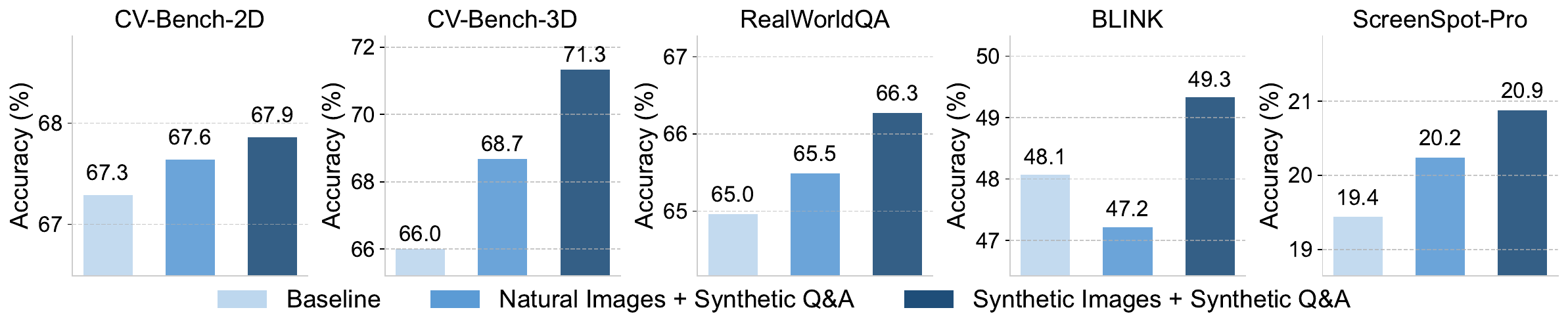}
    \caption{\textbf{Strict control experiment with matched synthetic QA.} Natural images and caption-conditioned synthetic images are compared under identical synthetic QA supervision and finetuning setup. Synthetic images achieve consistently stronger performance, with the largest gain on CV-Bench-3D.}
    \label{fig:natural_vs_synthetic_comparison_control}
\end{figure}

\paragraph{Findings and discussion.}
Figure~\ref{fig:natural_vs_synthetic_comparison_control} shows that, even under matched synthetic QA, the synthetic-image branch still performs better overall.
Compared with \emph{natural images + synthetic QA}, \emph{synthetic images + synthetic QA} improves CV-Bench-2D by +0.3 points, CV-Bench-3D by +2.6 points, RealWorldQA by +0.8 points, BLINK by +2.1 points, and ScreenSpot-Pro by +0.7 points.
The natural-image branch remains competitive and improves over the baseline on several metrics, but the synthetic-image branch provides consistently stronger performance, especially on visual perception evaluations (particularly CV-Bench-3D).
These findings further support that the synthetic process, particularly the caption-conditioned T2I reconstruction step, offers additional training value beyond caption-derived synthetic QA alone.

\section{T2I Model Ablations}
\label{app:opensource_t2i}

This section examines whether the gains from \dataengine depend on the proprietary T2I model used in the main pipeline, or whether a strong open-source image generator can provide similarly useful supervision.

\paragraph{Goal and protocol.}
The goal is to evaluate the reproducibility of the image-synthesis stage under the same downstream training budget and data construction recipe as in the preceding controlled ablations.
We again use Qwen2.5-VL-3B-Instruct~\citep{bai2025qwen25vl} as the backbone and keep the finetuning setup unchanged: unfreeze adapters, ViT, and LLM; use learning rates of $5\times10^{-6}$ for adapter/LLM and $5\times10^{-7}$ for ViT; train for one epoch with batch size 128.
The training task remains \emph{Orientation and Direction} with a matched final budget of 1k samples.
We compare three settings: (i) the no-finetuning baseline, (ii) \dataengine data synthesized with the open-source T2I model Qwen-Image-2512~\citep{wu2025qwenimagetechnicalreport}, and (iii) \dataengine data synthesized with the proprietary T2I model Gemini-2.5-Flash-Image~\citep{google2025gemini25flashimage,comanici2025gemini25} used in the main pipeline.
All non-T2I components, including QA construction, verification, and the downstream finetuning recipe, are kept fixed, so the comparison isolates the choice of image generator.

\begin{figure}[H]
    \centering
    \includegraphics[width=0.98\linewidth]{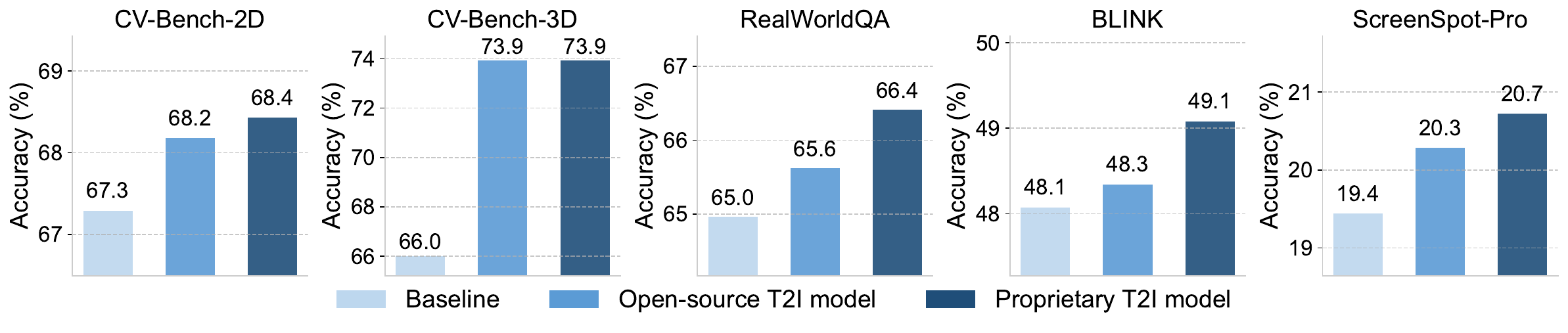}
    \caption{\textbf{Open-source vs.\ proprietary T2I models under matched 1k-data training.} Replacing the proprietary image generator in \dataengine with the open-source Qwen-Image-2512 still yields consistent improvements over the no-finetuning baseline. The proprietary T2I model remains slightly stronger overall, but the open-source variant retains most of the benefit, especially on visual perception benchmarks.}
    \label{fig:t2i_model_comparison}
\end{figure}

\paragraph{Results and discussion.}
Figure~\ref{fig:t2i_model_comparison} shows that the open-source T2I setting is already effective.
Compared with the baseline, finetuning on data synthesized by Qwen-Image-2512 improves CV-Bench-2D by +0.9 points, CV-Bench-3D by +7.9 points, RealWorldQA by +0.7 points, BLINK by +0.3 points, and ScreenSpot-Pro by +0.9 points.
The proprietary T2I variant remains the strongest overall: relative to the open-source branch, it improves CV-Bench-2D by +0.3 points, RealWorldQA by +0.8 points, BLINK by +0.7 points, and ScreenSpot-Pro by +0.4 points, while both variants achieve the same CV-Bench-3D score.
Thus, the open-source model does not fully match the proprietary frontier, but it still produces training data that consistently improves the downstream VLM.
These results support meaningful open-source reproducibility for \dataengine and suggest that its gains do not rely specifically on a proprietary T2I model; a strong open-source image generator is already sufficient to make the pipeline effective.

\section{Ablation on Verifier Choice}
\label{app:opensource_verifier}

One might view the verifier as imposing a hard upper bound on the downstream gains achievable by \dataengine. This section tests that interpretation directly.

\paragraph{Goal and protocol.}
The goal is to evaluate whether \dataengine mainly inherits supervision quality from a stronger verifier, or whether the pipeline remains effective even when the verifier is replaced by an open-source model that is not obviously stronger than the downstream baseline.
We again use Qwen2.5-VL-3B-Instruct~\citep{bai2025qwen25vl} as the backbone and keep the downstream finetuning setup unchanged: unfreeze adapters, ViT, and LLM; use learning rates of $5\times10^{-6}$ for adapter/LLM and $5\times10^{-7}$ for ViT; train for one epoch with batch size 128.
The training task remains \emph{Orientation and Direction} with a matched final budget of 1k samples.
To isolate the verifier choice, we fix the T2I model to the open-source Qwen-Image-2512~\citep{wu2025qwenimagetechnicalreport} used in the previous section and compare three settings: (i) the no-finetuning baseline, (ii) \dataengine data filtered by an open-source verifier instantiated as Qwen2.5-VL-3B-Instruct itself~\citep{bai2025qwen25vl}, and (iii) \dataengine data filtered by the proprietary verifier Gemini-3-Pro~\citep{google2025gemini3pro} used in the main pipeline.
All non-verifier components, including QA construction, T2I synthesis, and the downstream finetuning recipe, are kept fixed.

\begin{figure}[H]
    \centering
    \includegraphics[width=0.98\linewidth]{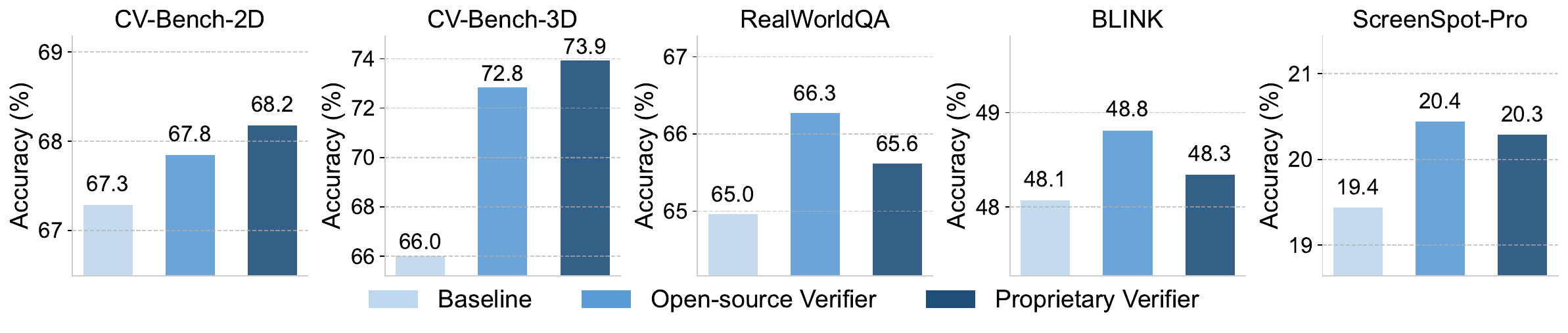}
    \caption{\textbf{Open-source vs.\ proprietary verifiers under matched 1k-data training.} With the T2I model fixed to Qwen-Image-2512, replacing Gemini-3-Pro with the open-source Qwen2.5-VL-3B-Instruct verifier still yields strong downstream gains. The two verifier choices trade wins across benchmarks rather than showing a one-sided dominance, suggesting that \dataengine's improvements are not simply inherited from the verifier ceiling.}
    \label{fig:verifier_comparison}
\end{figure}

\paragraph{Results and discussion.}
Figure~\ref{fig:verifier_comparison} shows that both verifier choices are effective and that the comparison is genuinely mixed.
Compared with the baseline, the open-source verifier improves CV-Bench-2D by +0.6 points, CV-Bench-3D by +6.8 points, RealWorldQA by +1.3 points, BLINK by +0.7 points, and ScreenSpot-Pro by +1.0 points.
The proprietary verifier also improves strongly over the baseline, with gains of +0.9 points on CV-Bench-2D, +7.9 points on CV-Bench-3D, +0.7 points on RealWorldQA, +0.3 points on BLINK, and +0.9 points on ScreenSpot-Pro.
Directly comparing the two verifier branches, the proprietary verifier is higher on CV-Bench-2D by +0.3 points and on CV-Bench-3D by +1.1 points, whereas the open-source verifier is higher on RealWorldQA by +0.7 points, BLINK by +0.5 points, and ScreenSpot-Pro by +0.2 points.
Hence, the proprietary verifier does not uniformly dominate, even though it is stronger in the main pipeline.
These results suggest that \dataengine is not simply obtaining useful supervision by delegating correctness to a stronger verifier alone; rather, the pipeline's task-aware generation-and-filtering structure itself contributes systematic training value.
They also demonstrate a practical degree of verifier-side generality and reproducibility: the framework remains effective with an open-source verifier, including one instantiated from the same model family as the downstream baseline.

\paragraph{Full-scale self-verification.}
We further scale the self-verifier setting from the matched 1k controlled study to all 10 tasks. We construct a 10K dataset using local Qwen-Image-2512 synthesis and Qwen2.5-VL-3B-Instruct itself as the verifier, and finetune the same Qwen2.5-VL-3B-Instruct backbone.
The resulting model improves over the pretrained baseline on all five visual-perception benchmarks, including +9.2 points on CV-Bench-3D and +5.4 points on $\mathrm{MMVP}_{\scriptstyle\text{pair}}$ (Table~\ref{tab:self_verifier_10k}).
Together with the matched verifier ablation above, this shows that useful \dataengine supervision does not require a stronger proprietary verifier.
The downstream model is Qwen2.5-VL-3B-Instruct with full-parameter SFT for one epoch. We use a global batch size of 128, learning rates of $5\times10^{-6}$ for the LLM/adapters and $5\times10^{-7}$ for the vision encoder, and the full 10K self-verifier dataset.

\begin{table}[H]
\centering
\small
\resizebox{0.99\linewidth}{!}{%
\begin{tabular}{lccccc}
\toprule
\textbf{Setting} & $\mathbf{MMVP}_{\scriptstyle\text{pair}}$ & $\mathbf{MMVP}_{\scriptstyle\text{single}}$ & \textbf{CV-Bench-2D} & \textbf{CV-Bench-3D} & \textbf{RealWorldQA} \\
\midrule
Qwen2.5-VL-3B baseline & 35.3 & 64.3 & 67.3 & 66.0 & 65.0 \\
Self-verifier 10K SFT & 40.7 & 68.7 & 69.3 & 75.2 & 65.9 \\
\midrule
Improvement & +5.4 & +4.4 & +2.0 & +9.2 & +0.9 \\
\bottomrule
\end{tabular}%
}
\caption{\textbf{Full-scale self-verification on visual perception benchmarks.} Performance of Qwen2.5-VL-3B-Instruct trained on the 10K self-verified \dataengine variant, where both image synthesis (Qwen-Image-2512) and verifier (Qwen2.5-VL-3B-Instruct) are performed using open-source models.}
\label{tab:self_verifier_10k}
\end{table}

\end{document}